\documentclass{article}

\PassOptionsToPackage{numbers, compress}{natbib}

\usepackage[preprint]{neurips_data_2021}

\usepackage[utf8]{inputenc} 
\usepackage[T1]{fontenc}    
\usepackage[ruled,vlined]{algorithm2e}
\usepackage{hyperref}       
\usepackage{url}            
\usepackage{booktabs}       
\usepackage{amsfonts}       
\usepackage{amsmath}
\usepackage{amssymb}
\usepackage{nicefrac}       
\usepackage{microtype}      
\usepackage{xcolor}         
\usepackage{dsfont}
\usepackage{graphicx}
\usepackage{enumitem}
\hypersetup{
    colorlinks=true,
    linkcolor=blue,
    filecolor=magenta,      
    urlcolor=blue,
}
\usepackage{subcaption}
\usepackage{listings}
\usepackage[frozencache=true,cachedir=.]{minted}

\newcommand{\blue}[1]{\textcolor{blue}{#1}}

\newcommand{\feat}{o}

\newcommand{\tar}{{\textnormal{target}}}
\newcommand{\mimax}{\textsc{MI-max}}
\newcommand{\dmin}{\textsc{D-min}}
\newcommand{\mizs}{MI(z,s)}
\newcommand{\entropyall}{H(s)}
\newcommand{\braxlines}{\textsc{Braxlines}}
\newcommand{\brax}{\textsc{Brax}}
\newcommand{\airl}{\texttt{AIRL}}
\newcommand{\gail}{\texttt{GAIL}}
\newcommand{\gcrl}{\texttt{GCRL}}
\newcommand{\diayn}{\texttt{DIAYN}}
\newcommand{\cdiayn}{\texttt{cDIAYN}}
\newcommand{\diaynfull}{\texttt{DIAYN\_FULL}}
\newcommand{\composer}{\textsc{Composer}}
\newcommand{\dnorm}[1]{\vert\vert#1\vert\vert}
\title{\textsc{Braxlines}: Fast and Interactive Toolkit for \\ RL-driven Behavior Engineering \\ beyond Reward Maximization}

\author{%
  Shixiang Shane Gu \\
  \texttt{shanegu@google.com} \\
   \And
   Manfred Diaz \\
  \texttt{diazcabm@mila.quebec} \\
   \AND
   Daniel C. Freeman \\
  \texttt{cdfreeman@google.com} \\
   \And
   Hiroki Furuta \\
   \texttt{furuta@weblab.t.u-tokyo.ac.jp} \\
   \And
   Seyed Kamyar Seyed Ghasemipour\\
   \texttt{kamyar@cs.toronto.edu} \\
   \And
   Anton Raichuk\\
   \texttt{raveman@google.com} \\
   \And
   Byron David\\
   \texttt{byrondavid@google.com} \\
   \AND
   Erik Frey \\
   \texttt{erikfrey@google.com}\\
   \And
   Erwin Coumans \\
   \texttt{erwincoumans@google.com}\\
   \And
   Olivier Bachem \\
   \texttt{bachem@google.com}\\
}

\begin{document}

\maketitle

\begin{abstract}
The goal of continuous control is to synthesize desired behaviors. In reinforcement learning (RL)-driven approaches, this is often accomplished through careful task reward engineering for efficient exploration and running an off-the-shelf RL algorithm. While reward maximization is at the core of RL, reward engineering is not the only -- sometimes nor the easiest -- way for specifying complex behaviors. In this paper, we introduce \braxlines, a toolkit for fast and interactive RL-driven behavior generation beyond simple reward maximization that includes \composer, a programmatic API for generating continuous control environments, and set of stable and well-tested baselines for two families of algorithms --mutual information maximization (\mimax) and divergence minimization (\dmin)-- supporting unsupervised skill learning and distribution sketching as other modes of behavior specification. In addition, we discuss how to standardize metrics for evaluating these algorithms, which can no longer rely on simple reward maximization. Our implementations build on a hardware-accelerated Brax simulator in Jax with minimal modifications, enabling behavior synthesis within minutes of training. We hope \braxlines~can serve as an interactive toolkit for rapid creation and testing of environments and behaviors, empowering explosions of future benchmark designs and new modes of RL-driven behavior generation and their algorithmic research. 

\end{abstract}

\section{Introduction}

 In the current era of neural networks and high-performance computing, large and well-engineered benchmarks~\citep{deng2009imagenet} have enabled key architectural and algorithmic breakthroughs, in computer vision~\citep{lecun2010mnist,cifar10,deng2009imagenet,alexnet,resnet,He_2017,Redmon_2016_CVPR}, natural language processing~\citep{transformer,wmt14_translate} and molecular biology~\citep{AlphaFold2021}. When scaled to the extreme levels of data and computing, these innovations can train models to master an impressive range of capabilities~\citep{devlin2019bert,radford2019language,brown2020language,radford2021learning,ramesh2021zeroshot,chen2021evaluating,bommasani2021opportunities} exhibiting \textit{developer-aware} generalization~\citep{chollet2019measure}, where the trained models can function even on datasets outside the developer's imagination. 
  Undoubtedly, the massive success of supervised learning (SL) has been largely due to large, diverse, and high-quality datasets, that are relatively simple to collect. In this paradigm, data generation essentially comes down to finding paired data: speech with texts~\citep{graves2013speech}, images with texts~\citep{deng2009imagenet,vinyals2015tell,VQA,ramesh2021zeroshot,radford2021learning}, words with words~\citep{luong2015effective,transformer,devlin2019bert,radford2019language,brown2020language,chen2021evaluating}, pixels with pixels~\citep{kingma2014autoencoding,goodfellow2014generative,rezende2016variational,oord2016pixel,kingma2017improving}. Moreover, a significant amount of this data can be mined automatically through the web~\citep{devlin2019bert,brown2020language,radford2021learning,chen2021evaluating} or collected relatively effortlessly with minor bottlenecks like manual human labeling~\citep{amazonmechanicalturk} or slow simulation~\citep{AlphaFold2021}.

 \begin{figure}[t]
    \centering
    \includegraphics[width=0.99\textwidth]{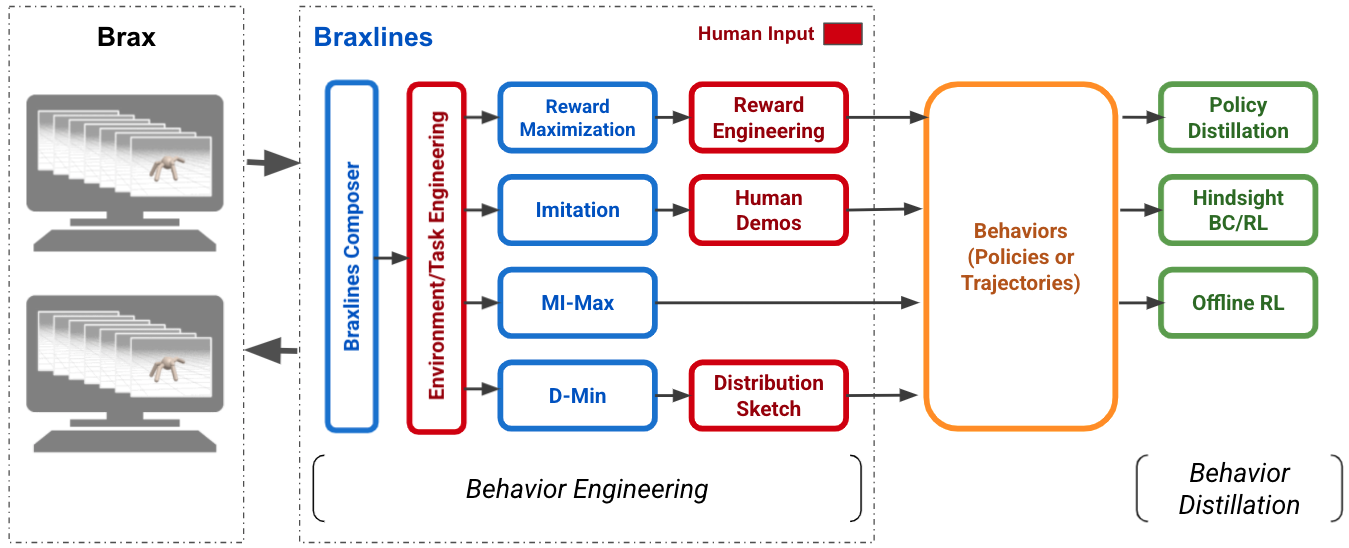}
    \caption{ \braxlines~vision: from behavior  \emph{engineering} to \emph{distillation}. Our \emph{behavior engineering} toolkit starts from \braxlines~\composer, a programmatic API that facilitates environment engineering and will accelerate the proliferation of novel environments for continuous control. Components in green represent multiple modes of RL-driven behavior engineering. Red components require human inputs and, except for environment engineering, require different modes of specifications from the designers (see \autoref{tab:alg_summary}). Importantly, all behaviors from these modes can be aggregated and fed to downstream behavior distillation modules. We believe that \textbf{maximizing simulation throughput} and \textbf{minimizing human inputs} as the keys to scalable behavior generation, where in this work we focus on enabling \mimax~and \dmin~as new modes of behavior engineering.}
    \label{fig:braxlines_vision}
\end{figure}

To mirror this success in RL, we may need efficient strategies for data generation. Data generation in RL~\citep{Sutton1998,mnih2015human,Lillicrap2016,gu2017deep} is significantly more costly (than in SL), the tasks design space is infinite, and in real-world environments like robotics, healthcare, or dialog modeling~\citep{gu2017deep,kalashnikov2018qtopt,murphy2001marginal,Jaques_2020} data is scarse. Even simulated environments~\citep{Bellemare2013ALE,gym2016openai,tassa2018deepmind,yu2019meta,cobbe2020leveraging,vinyals2017starcraft,savva2019habitat,gym_minigrid,dosovitskiy2017carla} are arguably far from having sufficient diversity, learnability, regularity, and downstream applicability to come close to enabling the levels of architectural breakthrough we have witnessed in SL.

Today, there exist data-centric techniques in RL such as policy distillation~\citep{rusu2016policy,hinton2015distilling,rusu2016progressive,levine2016endtoend}, hindsight imitation learning or RL~\citep{levine2016learning,lynch2019learning,andrychowicz2017hindsight,pong2018temporal,kalashnikov2018qtopt,jiang2019language,lynch2021language,chebotar2021actionable} and offline RL~\citep{levine2020offline,fujimoto2018bcq,kumar2020cql,fujimoto2021minimalist,ghasemipour2021emaq,Siegel2020KeepDW,matsushima2020deploy} that aim to learn faster by better utilizing prior policies or experiences. Behavior distillation techniques offer an important alternative to popular approaches~\citep{Bellemare2013ALE,gym2016openai,tassa2018deepmind,mnih2015human,schulman2015trpo,Schulman2017PPO,gu2016continuous, haarnoja2018sac,fujimoto2018td3} that often require to start \emph{tabula rasa} and rediscover meaningful behaviors over and over again. However, we argue that the question of how to obtain interesting and useful data for distillation methods is under-explored.
Prior work~\citep{fu2020d4rl} largely focuses on collecting human trajectories or obtaining trajectories from policies trained with engineered rewards. Both approaches are, practically and computationally, prohibitively expensive. 
In this paper, we focus on alternative methods to generate diverse and useful behaviors. 

We introduce \braxlines, an interactive toolkit for principled behavior creation for continuous control tasks (see \autoref{fig:braxlines_vision}) and claim the following key contributions:
\begin{itemize}[leftmargin=0.6cm,topsep=0pt,itemsep=0pt]

\item \braxlines~introduces a novel programmatic API for environment composition, \braxlines~\composer, that enables RL researchers to create continuous control environments from scratch or pre-defined components.
\item
\braxlines~provides clean, minimalistic implementations for two broad families of algorithms suitable for reward-free behavior engineering: mutual information maximization (\mimax) and divergence minimization (\dmin). They provide a rich set of alternative, sometimes complementary, algorithms for behavior engineering, including goal-conditioned RL~\citep{kaelbling1993learning,schaul2015uvf,andrychowicz2017hindsight,pong2018temporal}, empowerment maximization~\citep{Klyubin2005empowerment,eysenbach2019diayn,choi2021variational}, state marginal matching and adversarial inverse RL~\citep{ho2016generative,fu2018learning,ghasemipour2020divergence,lee2020efficient}. 
\item The accompanying toolkit provides absolute and stationary metrics for evaluating \mimax~and \dmin~algorithms, whose reward functions dynamically varies over training and across different implementation choices~\citep{Klyubin2005empowerment,eysenbach2019diayn,choi2021variational}. Some of these metrics essentially measure characteristics of behaviors themselves, such as how much control of degrees of freedom of the world an agent has, in a task-agnostic sense, providing more intrinsic insights into agent-environment interactions~\citep{varela1991embodied,brooks1991}.
\item The open-source implementation based on the Brax~\citep{brax2021github} physics library is extremely fast, allowing interactive creation of novel behaviors in a few minutes through Google Colab on a freely available 2$\times$2 with 8 cores TPU (see \autoref{tab:speed}). This effectively enables real-time interactive RL-driven environment and behavior designs.   
\end{itemize}

\begin{table}[t]
\small
\centering
    \begin{tabular}{|l|l|l|l|l|l|}
        \hline
        \hline
        \textbf{Library} & \textbf{Run-time} & \textbf{Steps} & \textbf{Resource} & \textbf{Simulator} & \textbf{Backends} \\
        \hline
        Baselines~\citep{baselines,zhao2021mutual} & 3.5 hours & $4.0 \times 10^7$ & 16 CPU + 1 GPU & MuJoCo & TensorFlow, OpenMPI \\
        \braxlines~& 3.5 min & $2.0\times10^8$ & 2$\times$2 TPU (free) & Brax & Jax \\
        \hline
        \hline
    \end{tabular}
\vspace{0.5pt}
\caption{Run-time speeds for \diayn~\citep{eysenbach2019diayn} on \emph{Ant} with different library and simulators. We adopt a recent open-source \diayn~ implementation released by~\citet{zhao2021mutual} built on Baselines~\citep{baselines}.}
\label{tab:speed}
\end{table}

\section{Related Work}
\label{sec:related_work}

\textbf{Benchmarks and Baselines}. Numerous benchmarks proposed over the years have spanned a wide range of tasks including simulated games of varying complexities~\citep{Bellemare2013ALE,vinyals2017starcraft}, continuous control and robotics \citep{todorov2012mujoco,tassa2018deepmind,yu2019meta,savva2019habitat,dosovitskiy2017carla}, or procedurally generated environments~\citep{cobbe2020leveraging,openendedlearningteam2021openended}. \braxlines~differentiate by focusing on improving the roundtrip time from conception to experimental results, increasing the interactivity of RL research. We leverage Brax~\cite{brax2021github} highly-paralellizable simulations to significantly reduce both time and computational requirements for experimentation in RL. Furthermore, using \braxlines~researchers can fix well-tested algorithmic baselines and iterate quickly over environment designs. Also, while many existing libraries~\citep{duan2016benchmarking,pong2018temporal,nair2018visual, baselines, Raffin_Stable_Baselines3_2020, liang2017rllib} provide fine-tuned collections of common deep RL~\citep{mnih2015human} algorithms, these works focus mostly on optimizing painstakingly hand-designed rewards (see \autoref{tab:reward_max} in Appendix \ref{sec:appendix_baselines}) using algorithms of varying complexity, from basic \cite{williams1992} to more advanced techniques \cite{Lillicrap2016, schulman2015trpo, Schulman2017PPO, fujimoto2018td3, haarnoja2018sac}. In contrast, \braxlines~provides generalized implementations for mutual information maximization (\textsc{MI-max}) \cite{choi2021variational} and divergence minimization (\textsc{D-min}) algorithms \cite{ghasemipour2020divergence}. While these approaches have been studied extensively \cite{Mohamed2015-bo, gregor2016variational, achiam2018variational, eysenbach2019diayn,sharma2020dynamics}, few benchmark implementations are available; often in isolated and not well-maintained repositories. To the best of our knowledge, we are the first to provide fast, testable, reproducible and minimalist baselines in this area (see \autoref{tab:info_max} in Appendix \ref{sec:appendix_baselines}). 

\textbf{Reward-free Behavior Engineering}. While arguably rewards are enough to produce intelligent behavior \cite{Silver2021-lz}, reward engineering \cite{dewey2014reward} bottlenecks the application of RL. Prior work have sought to address this issue by introducing other objectives such as human preferences elicitation \cite{hadfield2017inverse, christiano2017deep} or task-agnostic reward functions \cite{Schmidhuber2010creativityfun, oudeyerkaplan2007intrinsic, bellemare2016unifying, pathak2017curiosity}. On the latter, some recent works have emphasized empowerment \cite{Klyubin2005empowerment, salge2014empowerment, jung2011empowerment, Mohamed2015-bo} and skills/options discovery \cite{gregor2016variational, florensa2017stochastichrl, eysenbach2019diayn, wardefarley2019discern, hansen2020visr, sharma2020skillsrobotics} through \textsc{MI-max} objectives. \citet{choi2021variational} connects these two areas with a \emph{variational} approach to goal-conditioned RL \cite{kaelbling1993learning, andrychowicz2017hindsight, pong2018temporal, pong2020skew, wardefarley2019discern}. Section \ref{sec:mimax} discusses how \braxlines~leverages this variational approach to implement \mimax~algorithms. On the other hand, an alternative to generating reward-free behaviours is to learn from human demonstrations \cite{argall2009lfdsurvey}. The range of techniques has spanned from feature expectation matching \cite{ng2000inverserl, abbeel2004apprenticeship}, maximum entropy formulations \cite{ziebart2008maxentirl, wulfmeier2015maxentdirl}, to adversarial approaches to generative modelling \cite{ho2016generative, fu2018learning, finn2018adversarialirl}. Recently, unifying frameworks~\citep{ghasemipour2020divergence, ke2021fdivergenceil} emerged  under a probabilistic view, based on algorithms the minimization of \emph{f}-divergences \cite{renyi1961measures, csiszar1967divergence} between state-action and state marginal distributions. \braxlines~leverages, as we discuss in Section \ref{sec:dmin}, \citet{ghasemipour2020divergence} to implement algorithms of the \dmin~family.   

\textbf{Metrics for Reward-Free Behavior}. One limitation of reward-free approaches to RL has been the lack of quantitative evaluation metrics. Previous work in both \mimax~\citep{eysenbach2019diayn, achiam2018variational, sharma2020dynamics, sharma2020skillsrobotics} and \dmin~\citep{ghasemipour2020divergence, ke2021fdivergenceil, ho2016generative, fu2018learning, finn2018adversarialirl} families have relied on either qualitative analysis of the learned behavior, the value of the objective (which usually saturates), and the  performance on existing reward functions for downstream tasks. In the case of \mimax~approaches, as mutual information is difficult to scale to high-dimensional spaces \citep{poole2019vbmi, belghazi2018mine, oord2019representation}, it seems particularly hard to come up with quantitative metrics. \braxlines~toolkit provides a non-parametric, particle-based approximation to mutual information similar to PIC metrics~\citep{furuta2021pic}. We also leverage this approach to compute metrics for \dmin~algorithms.  Finally, we provide an implementation of Laten Goal Reaching (LGR) \citep{choi2021variational} that leverages the connection between goal-conditioned RL and \mimax~objectives. We discuss this metrics further in Section \ref{sec:metric} and contrast with prior evaluation techniques in Appendix~\ref{sec:prior_evaluation} and Table~\ref{tab:prior_evaluation}.

\textbf{Comparison to Other Benchmarks} We surveyed the  support of common RL algorithms in well-maintained \emph{open-source} repositories of baselines: Baselines~\citep{baselines}, Stable Baselines~\citep{Raffin_Stable_Baselines3_2020}, RLLab~\citep{duan2016benchmarking}, Tf-Agents~\citep{tfagents}, RLLib~\citep{liang2017rllib}, and PFRL / ChainerRL~\citep{fujita2021chainerrl}.  For each, we verified the support for continuous control, reward maximization algorithms like DDPG~\citep{Lillicrap2016}, A3C~\citep{mnih2016a2c}, PPO~\citep{Schulman2017PPO}, TRPO~\citep{schulman2015trpo}, TD3~\cite{fujimoto2018td3}, SAC~\cite{haarnoja2018sac} and ES~\citep{salimans2017evolution}.
Similarly, we investigated whether existing baselines provide support for algorithms of the \mimax~and \dmin~families. With the exception of RLLab and OpenAI Baselines, there were no other benchmark implementation for \mimax~\gcrl~\citep{kaelbling1993learning}, \diayn~\citep{eysenbach2019diayn}, DISCERN~\citep{wardefarley2019discern}, VISR~\citep{hansen2020visr} or \dmin~algorithms~\citep{ho2016generative, ghasemipour2020divergence, ke2021fdivergenceil}. \autoref{tab:reward_max} and \ref{tab:info_max} in Appendix \ref{sec:appendix_baselines} show the findings of our survey. \braxlines~is, \textbf{to the best of our knowledge, the first reproducible and reusable open-source library of baseline implementations for reward-free RL algorithms}. 

\section{Background} 
\label{sec:algorithm}

\textbf{Notations} We consider a Markov Decision Process (MDP) to be a tuple characterized by $(\mathcal{S}, \mathcal{A}, p, r, \rho_0, \gamma)$; state space $\mathcal{S}$, action space $\mathcal{A}$, transition probability $p: \mathcal{S} \times \mathcal{A} \rightarrow \mathcal{S}$, reward function $r: \mathcal{S} \times \mathcal{A} \rightarrow \mathbb{R}$, initial state distribution $\rho_0$, and discount factor $\gamma \in (0,1]$.
 A policy is a function $\pi: \mathcal{S} \rightarrow \mathcal{A}$ that defines behaviors.
In this work, we denote the state-action and marginal state distribution of a policy as $\rho^{\pi}(s,a)$ and $\rho^{\pi}(s)$, respectively.

\subsection{Mutual Information Maximization (\mimax)}
\label{sec:mimax}

Reward-free RL typically considers the maximization of mutual information between actions and future states. Objectives such as empowerment~\citep{Klyubin2005empowerment, Mohamed2015-bo} or unsupervised skills discovery~\citep{gregor2016variational, eysenbach2019diayn, achiam2018variational, wardefarley2019discern} propose the maximization of mutual information between states and abstracted actions $z \sim p(z)$ (commonly known as \emph{skills}~\citep{ravindran2002hierarchical}). Algorithms in this family learn policies $\pi(a|s, z)$ conditioned on this latent actions. Recently, \citet{choi2021variational} presented Variational Goal-Conditioned RL (V-\gcrl), a framework that unifies Goal-Conditioned RL (\gcrl)~\citep{kaelbling1993learning} and a variational approach to \mimax, derived from the mutual information maximization objective:
\begin{align}
    \max_\pi \text{MI}(z, s ; \pi) &= \max_\pi H(s ; \pi) - H(s|z ; \pi) \label{eq:mi_s} \\
    &= \max_\pi H(z) - H(z| s; \pi) \nonumber\\
    &= \max_\pi \iint p(z) \rho^\pi(s|z) \log p^\pi(z|s) - \int p(z) \log p(z) \label{eq:mi_z}
\end{align}
where $MI(z, s; \pi)$ is the mutual information between the distribution of latent actions or skills $z \sim p(z)$ and the state marginal distribution $s \sim \rho^\pi(s)$, under $z$-conditioned policy $\pi(a|s, z)$.
The objective in Eq.~\ref{eq:mi_s} is a \textit{generative} formulation while Eq.~\ref{eq:mi_z} defines a \textit{discriminative} of the \mimax~objective.
Due to the intractability of the true posterior $p^\pi(z|s)$, a variational bound of Eq.~\ref{eq:mi_z} is used instead: 
\begin{align}
    \max_\pi \text{MI}(z, s ; \pi) &\geq \max_\pi \iint p(z) \rho^\pi(s|z) \log q(z|s) - \int p(z) \log p(z) \label{eq:vmi_z}
\end{align}
that enables most of the algorithms in the \mimax~family~\citep{kaelbling1993learning,li2017infogail,eysenbach2019diayn,wardefarley2019discern,hansen2020visr, sharma2020dynamics}. 

\subsection{Divergence Minimization (\dmin)}
\label{sec:dmin}
    
    Inverse RL (IRL)~\citep{ng2000inverserl,abbeel2004apprenticeship} aims to infer the reward function and subsequently obtain an optimal policy from an expert behavior.
    Seminal work~\citep{ho2016generative} has demonstrated that the problem of MaxEnt IRL is equivalent to matching the state-action marginal of a policy, $\rho^\pi(s,a)$, to that of the expert's, $\rho^{\textnormal{exp}}(s,a)$. Thus, distribution-matching GAN~\citep{goodfellow2014generative} techniques can be used for IRL with little expert data, leading to Adversarial Imitation Learning (AIL) of methods~\citep{ho2016generative,fu2018learning}. Recently, the work in~\citep{ghasemipour2020divergence, ke2021fdivergenceil} have shown that  to correspond to different choice of $f$-divergences for state-action distribution matching. 
    
    An interesting modification studied in~\citep{ghasemipour2020divergence,lee2020efficient,hazan2019provably} matches $\rho^\pi(s)$ to a desired distribution $\rho^\tar(s)$ and allows AIL to match behavior not necessarily generated by an expert policy, and can even be designed without any access to the underlying MDP. 
    For a given choice of $f$-divergence, minimizing $D_f$ of state marginals can be accomplished through an adversarial optimization setup \citep{ghasemipour2020divergence}.
   \begin{align}
        \min_\pi D_f(\rho^\pi(s), \rho^{\textnormal{target}}(s)) =    \max_\pi \mathds{E}_{\rho^\pi(s)}\Big[f^*(T_\omega(s))\Big] =  \max_\pi \mathds{E}_{\rho^\pi(s)}\Big[r^*(s)\Big]
   \end{align}
    recovering optimization objectives similar to \gail \citep{ho2016generative} (Jensen-Shannon divergence), \airl \citep{fu2018learning} (reverse KL divergence), and \texttt{F}-\airl \citep{ghasemipour2020divergence} (forward KL divergence). 
    
    In these settings, the discriminator is a binary classifier distinguishing between target states and states visited by the policy, trained using cross entropy. The reward functions for training the policy are obtained from:
        \begin{align}
            r^{\text{\gail}}(s) &:= \text{log }D(s) = \log \rho^\tar(s) - \log\left(\rho^\tar(s) + \rho^\pi(s) \right) \label{eq:gail} \\
            h(s) &:= \text{log }D(s) - \text{log }(1 - D(s)) \nonumber\\
            r^{\text{\airl}}(s) &:= h(s) \label{eq:airl}  =\log \rho^\tar(s) - \log \rho^\pi(s) \\
            r^{\text{\texttt{MLE}}}(s) &:= \log \rho^\tar(s) \label{eq:mle}.
        \end{align}
    where $D(s)$ is the probability the discriminator assigns for $s$ being from the target distribution. For a more in depth discussion, we refer the reader to \citep{ghasemipour2020divergence, ke2021fdivergenceil}\footnote{The last reward Eq.~\ref{eq:mle} is technically not an instance of \dmin~algorithms and is a plain stationary reward function, but it is effectively \airl (Eq.~\ref{eq:airl}) without $H^\pi(s)$ term. In~\autoref{fig:d_min}, we show that sometimes this presents competitive results for matching multi-modal distribution as a full \dmin~algorithm.}.
    
\section{Tools for Behavior Engineering}
\label{sec:metric}

\braxlines~for behavior engineering is built over three pillars: (1) an environments composer, (2) a set of stable baselines, and (3) a collection of metrics for reward-free behavior evaluation.

\subsection{\braxlines~\composer}

\composer~ is designed for \textbf{modularity} and \textbf{reusability}. \autoref{fig:composer} illustrates examples of environments that can be composed easily using \composer. See Appendix~\ref{sec:composer_api_example} for an Ant Push task example constructed under 50 lines. Since all observations, reward functions, and scene components can be \textit{combined} and \textit{reused}, \composer~allows programmatic procedural generation of parameterized environments. Combined with \brax's PPO/ES methods and \braxlines' \mimax/\dmin~implementations that allow training \textit{within minutes}, these enable designers to quickly design, debug, and tune tasks.

\begin{figure}[h]
    \centering
    \begin{subfigure}{0.3\textwidth}
        \centering
        \includegraphics[width=\textwidth]{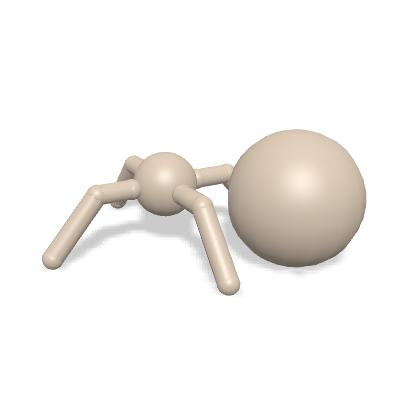}
    \end{subfigure}
    \begin{subfigure}{0.3\textwidth}
        \centering
        \includegraphics[width=\textwidth]{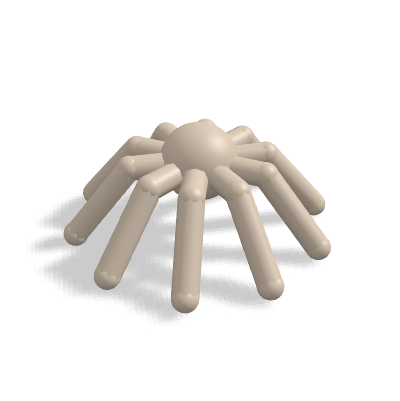}
    \end{subfigure}
    \begin{subfigure}{0.3\textwidth}
        \centering
        \includegraphics[width=\textwidth]{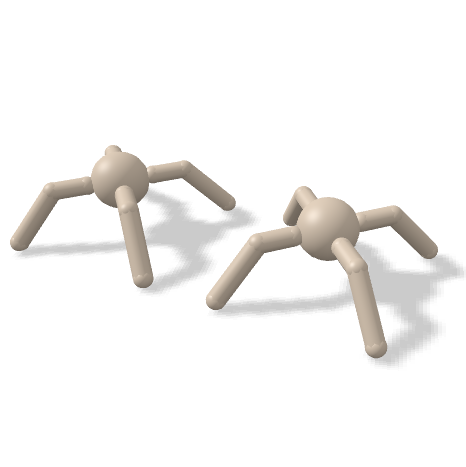}
    \end{subfigure}
   \caption{\braxlines~\composer~API is a flexible and fast environment engineering tool for continuous control. The space of environments designs spans from (left) goal-based locomotion + manipulation task, (middle) parameterized morphologies, to (right) multi-agent systems. See Appendix \ref{sec:composer_api_example} for an example on how to construct Ant Push (left) in under 50 lines of code.}
    \label{fig:composer}
\end{figure}

\subsection{Stable Baselines for Reward-free Behavior Generation}

\braxlines~is designed to optimize for \textbf{speed} and \textbf{minimalism} of algorithms in the \mimax~and \dmin~families. Prior implementations~\citep{eysenbach2019diayn,ho2016generative,fu2018learning,ghasemipour2020divergence,lee2020efficient,nachum2018hiro} rely on heavily modifying different RL optimizers (e.g. SAC \cite{haarnoja2018sac}, TD3 \cite{fujimoto2018td3}, PPO \cite{Schulman2017PPO}, MPO \cite{abdolmaleki2018mpo}), making comparative analyses and code reuses difficult. The benchmark implementations we provide are unified under Brax's PPO~\citep{brax2021github}, enabling stable and \textit{speedy} training on medium-size problems within minutes. \autoref{tab:speed} shows \braxlines~speed gains of orders of magnitude compared to existing \diayn~\cite{eysenbach2019diayn} baselines. Furtheremore, we leverage recent unification papers~\citep{ghasemipour2020divergence, choi2021variational} to keep \braxlines~design as clean and \emph{minimal} as possible. A minimalistic approach to RL~\citep{fujimoto2021minimalist} ensures that codebases does not rely on extensive hyperparameter designs~\citep{henderson2018deep,andrychowicz2021what,engstrom2019implementation,furuta2021coadapt}, and can be debugged, reproduced, extended, and mixed easily downstream.

\begin{table}
\small
\centering
    \begin{tabular}{|l|c|l|l|}
        \hline
        \hline
        \textbf{Algorithm} & \textbf{Family} & \textbf{Implementation} & \textbf{User Specification} \\
        \hline
        \gcrl~\citep{kaelbling1993learning,andrychowicz2017hindsight,pong2018temporal, choi2021variational} &\mimax & Fix $q(z|s)= \mathcal{N}(\feat(s),\sigma I)$, $p(z)$ + Offset & $p(z)$, $\feat(\cdot)$ \\
        V-\gcrl~\citep{choi2021variational,eysenbach2019diayn,wardefarley2019discern,hansen2020visr} & \mimax& Param $q_\theta(z|\feat(s))$+ Fix $p(z)$ + SN + Offset  & $\feat(\cdot)$\\
        \hline
        \hline
        \gail~\citep{ho2016generative} &\dmin & Eq.~\ref{eq:gail} + Offset & $\feat(\cdot)$, $\rho^\tar(\feat(s))$ \\
        \airl~\citep{fu2018learning} & \dmin&Eq.~\ref{eq:airl} + Offset & $\feat(\cdot)$, $\rho^\tar(\feat(s))$ \\
        \hline
        \hline
    \end{tabular}
\vspace{0.5pt}
\caption{A summary of  example \braxlines~algorithms explained in terms of the \textbf{Algorithm Family}/\textbf{Implementation} decomposition in~\cite{furuta2021coadapt}. \braxlines~emphasizes on minimal implementations with little to no code-level optimization (only spectral normalization (``SN'') \cite{miyato2018spectral,choi2021variational} on discriminator $q_{\theta}(z|\cdot)$) and adding positive constant offsets (``Offset''), equivalent survival bonus, to ensure that rewards almost always positive (see Appendix~\ref{sec:hyperparameters}). \textbf{User Specification} lists what \braxlines~users are required to define to perform feature engineering $\feat(\cdot)$, and marginal state distribution matching $\rho^\tar(\cdot)$. Importantly, since V-\gcrl~has representation learning capacity, $p(z)$ can often be simply a fixed uninformative prior, e.g. zero-mean Gaussian or a uniform Categorial~\citep{eysenbach2019diayn,choi2021variational}.}
\label{tab:alg_summary}
\end{table}

We present in \autoref{tab:alg_summary} an analysis of the flexibility of \braxlines~using an \emph{algorithm-implementation} decomposition similar to~\citet{furuta2021coadapt} that separates \textbf{algorithm} (i.e., mathematical and algorithmic choices) from \textbf{implementation} details (i.e., implementation and code-level optimizations). In addition, we define what are users required to specify to turn a family of algorithms into a concrete instance (e.g., \mimax~to \diayn). In the sections that follow, we present more detailed explanations of both \mimax~and \dmin~families, and how are they supported in \braxlines. We support flexible parametrizations of $q(z|s)$ that could take the form of an isotropic Gaussian distribution $q(z|s)=\mathcal{N}(O; \mu, \sigma^2 I)$,or a parametric tractable posterior $q_{\theta}(z|s)$ learned using function approximation (i.e., neural networks). Similarly, \braxlines~provides baseline implementations for variations of the AIL family of algorithms. In the case of \dmin~ algorithms, we provide the flexibility for users to specify feature engineering functions as well as behavior distributions over this feature space. Together, algorithms in both families are a key ingredient  for generating diverse behaviors through little human intervention.

\subsection{Metrics}

Both \mimax~and \dmin~algorithms infer dynamic reward functions during learning (except in few special cases like \gcrl), which can change arbitrarily across training iterations, as well as across different implementation and hyperparameter choices (e.g. discrete or continuous $z$-space in \mimax, or different choices of reward transformation in \dmin~as in \autoref{tab:alg_summary}). This makes episodic rewards inapplicable for quantitatively evaluating optimization convergence or final performances. Common alternatives are also limited and hard to automate: \mimax~algorithms often use qualitative visual inspections or down-stream task performances~\citep{eysenbach2019diayn,sharma2020dynamics,sharma2020skillsrobotics}, while \dmin~algorithms mostly focus on imitation learning and frequently use task rewards directly to compare how close to the experts they got~\citep{ho2016generative,fu2018learning,ghasemipour2020divergence}. 

Inspired by prior works~\citep{choi2021variational}, we provide objective metrics for \mimax~and \dmin, that (1) provide an intuitive and interpretable \textbf{absolute units} of performance measure applicable to any algorithm instance within the same family, and (2) are \textbf{stationary} throughout learning, meaning that they do not depend of quantities that change over the course of training or across runs of an experiment (such as rewards given by the discriminator).

\subsubsection{\mimax~Metrics}

We implemented two metrics for evaluating the behaviors generated by algorithms in the \mimax~ family that leverage approximations to mutual information. These metrics compute, in general, how much an agent learned to control each dimension of the environment.   
\paragraph{Particle-based Mutual Information Approximation.} We can directly adopt similar techniques in~\citet{furuta2021pic} to estimate empowerment of a trained policy. It uses straight-forward discretization for tractable non-parametric estimation in 1D and 2D. Intuitively this metric quantifies \textbf{how much predictive control over each dimension(s) of the environment an agent has learned}, a direct measure of the original \mimax~objective in Eq.~\ref{eq:mi_s} (without variational approximation). Note that reliable mutual information estimation in high dimensions is still an actively researched problem,
but some scalable estimation techniques~\citep{poole2019vbmi,belghazi2018mine,oord2019representation} will be added later.

\begin{algorithm}[H]
\SetAlgoLined
\KwIn{agent $\pi(a|s, z)$; intent $p(z)$; feature function $\feat(s)$; bin size $B$; bin range $(a,b)$ } 
\KwOut{Mutual information estimate $\hat{\text{MI}}$} 
  \For{$n=1,...,N$}{
    Sample $z_n\sim p(z)$  \blue{ // Sample an agent intent}\\
    Sample $s_{n,m,t}\sim\pi(\cdot|\cdot,z_n)$ \blue{// Collect $M$ episodes of horizon $T$ for $z_n$} \\
    \blue{// Estimate conditional entropy in $\feat$-space for $z_n$ using np.histogram(2d)($B, (a, b)$)} \\
    Estimate $h_n$ with $p(\feat|z_n)\approx \frac{1}{MT}\sum_{m,t}\delta(\feat=\feat(s_{n,m,t}))$ \\
  }
    \blue{// Estimate marginal entropy in $\feat$-space for $z_n$ using np.histogram(2d)($B, (a, b)$)} \\
    Estimate $h$ with $p(\feat)\approx \frac{1}{NMT}\sum_{n,m,t}\delta(\feat=\feat(s_{n,m,t}))$ \\
  $\hat{\text{MI}} =h - \frac{1}{N}(\sum_n h_n)$   \blue{// Compute MI estimate }\\
 \caption{Particle-based Mutual Information Approximation}
 \label{alg:mi}
\end{algorithm}

\textbf{Latent Goal Reaching.}~ In our experiments, we fixed $N=1$ for Algorithm~\ref{alg:lgr} and use the deterministic sampling (for Gaussians, take the mean; for Categoricals, take the argmax). LGR for \gcrl~ in Section~\ref{sec:mimax} corresponds exactly to the standard goal-reaching reward evaluation. 
While the LGR metric in Algorithm~\ref{alg:lgr} (see Appendix~\ref{sec:lgr}) is motivated intuitively as a task-oriented metric from \gcrl~\citep{choi2021variational} and appears unrelated to the \mimax~objective (Eq.~\ref{eq:mi_z}) or its approximation (Algorithm~\ref{alg:mi}), it can also be viewed as a crude approximation to MI, as we show in Appendix~\ref{sec:lgr_deriv}. 

\subsubsection{\dmin~Metrics}

\textbf{Particle-based Divergence Approximation.}~
 For lower dimension cases, we use the energy distance for non-parametric estimation. Given two distributions $p$ and $q$, the energy distance is defined as,
    \begin{align}
        d(p, q) := 2\mathds{E}_{x \sim p, y \sim q}\vert\vert x - y \vert\vert - \mathds{E}_{x, x' \sim p}\vert\vert x - x' \vert\vert - \mathds{E}_{y, y' \sim q}\vert\vert y - y' \vert\vert\nonumber
    \end{align}
 where $\vert\vert\cdot\vert\vert$ is the Euclidean norm but may be any desired metric.
 For higher dimensional and more complex distributions, adhoc notions of distance (such as Frechet Inception Distance \citep{heusel2017gans}) have been proposed.

\begin{algorithm}[H]
\SetAlgoLined
\KwIn{agent $\pi(a|s)$; $\feat(s)$; target samples $\feat^\tar_{1:L}=\feat(s^\tar_{1:L})\sim p^\tar(s)$ }
\KwOut{Energy distance estimate $\hat{\text{D}}$} 
    Sample $\feat_{m,t} = \feat(s_{m,t}), s_{m,t}\sim\pi$ in $\mu$ \blue{// Collect $M$ episodes of horizon $T$}  \\
  \blue{// Compute energy distance} \\
  $\hat{\text{D}} = \frac{2}{MTL}\sum_{l,m,t} \dnorm{\feat^\tar_l- \feat_{m,t}} - \frac{1}{M^2T^2}\sum_{m,t,m',t'} \dnorm{\feat_{m,t}-\feat_{m',t'}} -  \frac{1}{L^2}\sum_{l,l'} \dnorm{\feat^\tar_l- \feat^\tar_{l'}}$ \\
 \caption{Energy Distance}
  \label{alg:energy}
\end{algorithm} 

\begin{figure*}[t]
\centering
\includegraphics[width=\textwidth]{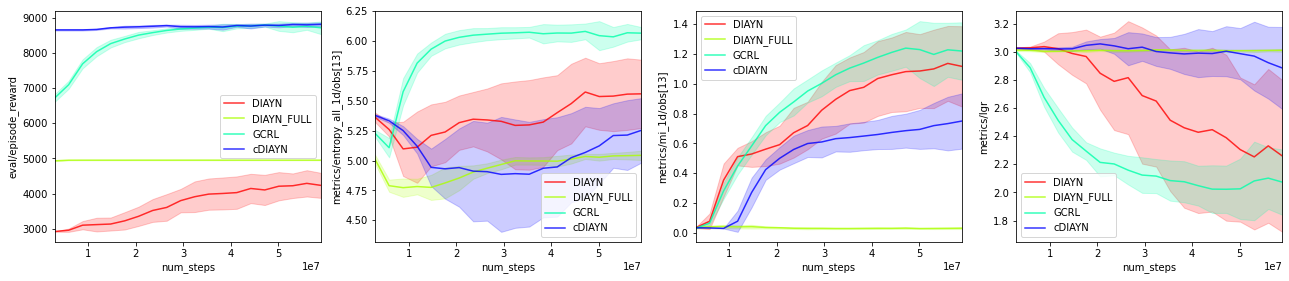}

\vskip -0.05in
\caption[Caption without FN]{
\mimax~results on Ant averaged over 10 seeds: (from left to right) (a) episodic discriminator reward, (b, c) MI and $H(s)$ estimates from Algorithm~\ref{alg:mi}\footnotemark, and (d) LGR from Algorithm~\ref{alg:lgr}. Since the discriminator reward scale (a) can vary arbitrarily among different instances of \mimax~algorithms, it is not useful for comparisons. Conversely, (b-d) metrics have the same units for any \mimax~algorithm, allowing comparisons.
\label{fig:mi_max}}
\end{figure*}

\section{Experiments}
\label{sec:exp}

We provide both illustrative Google Colab examples \url{https://github.com/google/brax/tree/main/notebooks/braxlines} for interactive environment composition and training, and a set of quantitative benchmark results that could accelerate future algorithmic research in \mimax~and \dmin~RL. 
\footnotetext{XY velocities correspond to dimensions $(13, 14)$ of Ant env. Since each dimension had similar MI estimates, we only show for MI and entropies for dim$=(13,)$. See Appendix~\ref{sec:hyperparameters} for more discussions.}
Importantly, due to two orders of training speedups as measured in Table~\ref{tab:speed}, a medium-size environment can be composed and trained in a few minutes\footnote{If you enable dense evaluations of the quantitative metrics in Section~\ref{sec:metric}, it could slow down to 10-20 minutes, but we mainly recommend these during initial debugging or full quantitative benchmarking.} on Google Colab with 2$\times$2, 8 cores TPU, enabling \textit{interactive} environment designs and algorithm deployments.
All \braxlines~codes, documentation, result videos, and Colab examples are accessible at
\url{https://github.com/google/brax/tree/main/brax/experimental/braxlines}. Experimental details and hyperparameters, additional benchmark results, and ablation studies are available in Appendix~\ref{sec:appendix_baselines}.

\begin{table}
\small
\centering
    \begin{tabular}{|l|c|c|c|c|}
        \hline
        \hline
        Environment & \mimax~algo.& MI$(s,z)$ & $H(s)$ & -LGR \\
        \hline
        HalfCheetah & \diayn & $ \mathbf{1.815 \pm 0.201} $ & $ 5.164 \pm 0.455 $ & $ -0.492 \pm 0.154 $\\
HalfCheetah & \diaynfull & $ 0.490 \pm 0.156 $ & $ \mathbf{5.519 \pm 0.178} $ & $ -1.397 \pm 0.287 $\\
HalfCheetah & \gcrl & $ 1.626 \pm 0.157 $ & $ 5.307 \pm 0.173 $ & $ \mathbf{-0.293 \pm 0.356} $\\
HalfCheetah & \cdiayn & $ 1.551 \pm 0.098 $ & $ 5.104 \pm 0.353 $ & $ -0.741 \pm 0.365 $\\
        \hline
Ant & \diayn & $ 1.115 \pm 0.268 $ & $ 5.558 \pm 0.286 $ & $ -2.260 \pm 0.539 $\\
Ant & \diaynfull & $ 0.034 \pm 0.006 $ & $ 5.044 \pm 0.039 $ & $ -3.010 \pm 0.010 $\\
Ant & \gcrl & $ \mathbf{1.218 \pm 0.192} $ & $ \mathbf{6.066 \pm 0.051} $ & $ \mathbf{-2.074 \pm 0.232} $\\
Ant & \cdiayn & $ 0.750 \pm 0.185 $ & $ 5.253 \pm 0.270 $ & $ -2.885 \pm 0.291 $\\
\hline
\hline
Humanoid& \diayn & $ \mathbf{0.927 \pm 0.128} $ & $ \mathbf{6.074 \pm 0.088} $ & $ \mathbf{-2.313 \pm 0.266} $\\
Humanoid& \diaynfull & $ 0.071 \pm 0.007 $ & $ 5.676 \pm 0.081 $ & $ -3.051 \pm 0.028 $\\
Humanoid& \gcrl & $ 0.769 \pm 0.146 $ & $ 6.058 \pm 0.063 $ & $ -2.755 \pm 0.115 $\\
Humanoid& \cdiayn & $ 0.297 \pm 0.147 $ & $ 5.677 \pm 0.060 $ & $ -2.738 \pm 0.112 $\\
        \hline
        \hline
    \end{tabular}
\vspace{0.5pt}
\caption{Benchmark for \mimax~algorithms averaged over 10 seeds.}
\label{tab:mimax_results}
\end{table}

\begin{table}[ht]
\small
\centering
    \begin{tabular}{|l|c|c|c|c|}
        \hline
        \hline
        Environment & \dmin~algo.& -Energy Distance \\
        \hline
        HalfCheetah & \airl & $ -1.448 \pm 0.843 $\\
HalfCheetah & \gail & $ \mathbf{-0.516 \pm 0.640} $\\
HalfCheetah & \gail2 & $ -2.488 \pm 0.355 $\\
HalfCheetah & \texttt{MLE} & $ -2.614 \pm 0.415 $\\
        \hline
       Ant & \airl & $ -2.588 \pm 1.194 $\\
Ant & \gail & $ \mathbf{-0.802 \pm 0.460} $\\
Ant & \gail2 & $ -2.898 \pm 0.946 $\\
Ant & \texttt{MLE} & $ -1.829 \pm 0.573 $\\ 
        \hline
        Humanoid& \airl & $ \mathbf{-1.122 \pm 0.210} $\\
Humanoid& \gail & $ -1.790 \pm 0.413 $\\
Humanoid& \gail2 & $ -1.471 \pm 0.528 $\\
Humanoid& \texttt{MLE} & $ -2.037 \pm 0.973 $\\
        \hline
        \hline
    \end{tabular}
\vspace{0.5pt}
\caption{Benchmark for \dmin~algorithms averaged over 10 seeds.}
\label{tab:dmin}
\end{table}

\textbf{\mimax~} \autoref{fig:mi_max} shows the training curves with respect to (a) inferred reward, (b, c) mutual information MI and state entropy $H(s)$ estimates, and (d) LGR estimate.
The scales of reward between \cdiayn~ (continuous \diayn)/\gcrl~ and \diayn~ are very different due to the continuous and discrete $z$ respectively, and they also vary non-monotonically during training due to the interplay between RL and discriminator fitting.  
In contrast, both MI and LGR estimates have the same comparable absolute units for any algorithm in the family, making them possible for estimating the effectiveness of learning and compare algorithmic performances.
Additional results are presented in~\autoref{tab:mimax_results} and Appendix~\ref{sec:additional_results}.

\begin{figure*}[t]
\centering
     \begin{subfigure}[]{0.45\textwidth}
        \centering
         \includegraphics[width=0.99\linewidth]{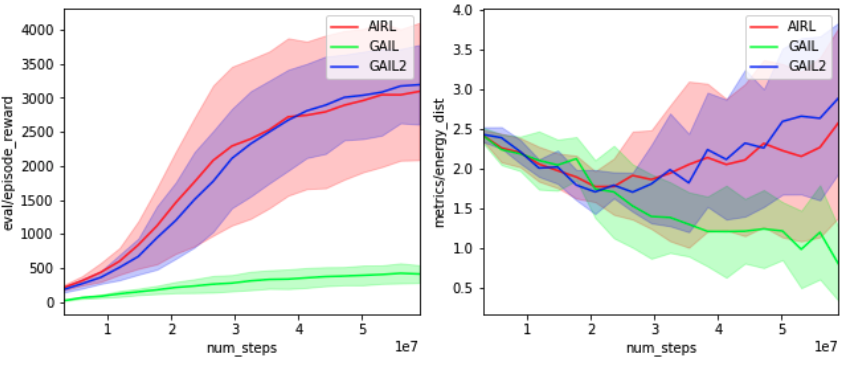}
    \end{subfigure}
         \begin{subfigure}[]{0.45\textwidth}
        \centering
          \includegraphics[width=0.99\linewidth]{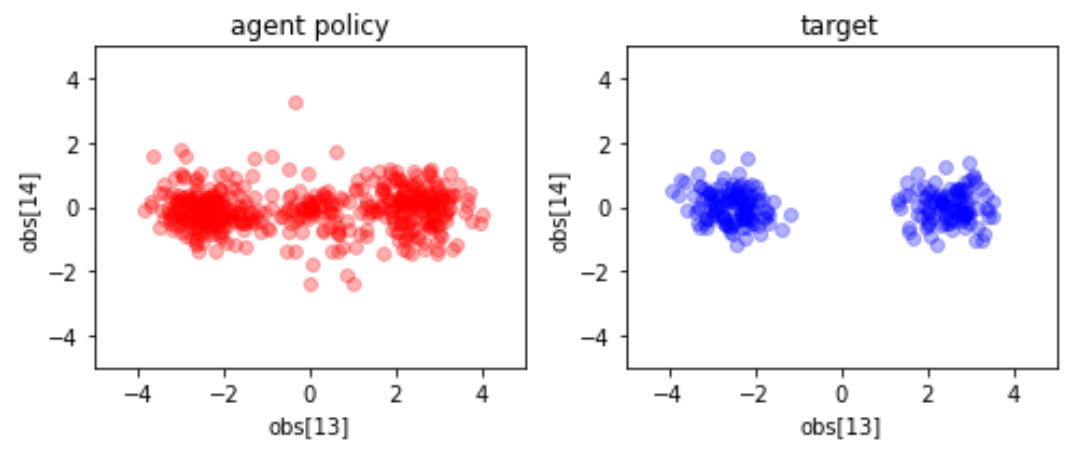}
    \end{subfigure}
\vskip -0.05in
\caption{\dmin~results for Ant averaged over 10 seeds: (from left to right) (a, b) episode reward and energy distance metric across training iterations; (c, d) visualizations of learned policy state marginal distribution and the target. Similarly to \autoref{fig:mi_max} of \mimax, (a) shows that episode rewards cannot be used as a metric for evaluating learning progress within each run or across runs of different choices. Our proposed metric (b), however, can successfully measure the convergence and algorithm performances (\gail~has lowest energy distance and performs the best). (c) shows the result of matching a bi-modal distribution in XY velocity space, where the Ant acquired a hopping behavior.\label{fig:d_min}}
\end{figure*}

\textbf{\dmin~} Similarly to~\autoref{fig:mi_max}, \autoref{fig:d_min} (left) presents the training curves for reward and divergence estimate.
Unsurprisingly, while the reward discriminator loss curves are incomparable, the energy distance estimate curve can be reliably used to measure convergence and compare which algorithmic variant performs the best in terms of state-marginal matching. 
Additional results are presented in~\autoref{tab:dmin} and Appendix~\ref{sec:additional_results}.

\textbf{Limitations}
\begin{itemize}
    \item \textbf{Feature Engineering and Metric Space Assumption}: Both \mimax~and \dmin~objectives indirectly involve density estimation, which is sensitive to high dimensionality and poor conditioning. An implicit assumption in all the above metrics is that $o(s)$ by the user specifies a low-dimensional, good metric space where Euclidean distance is a good measure for proximity. Since our emphasis is on behavior generation in simulation with access to simulator states (which could be used by downstream behavior distillation procedures~\citep{levine2016endtoend,lynch2019learning}), this remains a reasonable assumption in many cases. 
    \item \textbf{Sample Efficiency}: Sample efficiencies are not the primary concern for \braxlines, as the code is designed to maximize the benefits of hardware-accelerated simulations and serve as an interactive toolkit for behavior designers and RL researchers exploring high-level concepts. While these are not tuned with respect to sample efficiency compared to many prior works~\citep{sharma2020skillsrobotics}, they exhibit orders of magnitudes speedups in training and often better final performances (e.g. \diayn~on Humanoid has not been successful~\citep{eysenbach2019diayn,sharma2020dynamics,choi2021variational}).
\end{itemize}

\section{Conclusion}
In this paper, we introduced \braxlines, a fast and interactive toolkit for behavior synthesis beyond reward engineering in continuous control that unlocks four key bottlenecks in RL research: (i) fast environment generation, (ii) difficulty and limitations of reward engineering, (iii) slow iteration speeds, and (iv) lack of metrics on data or tasks properties.
Our experimental analysis showcased how two families of algorithms – mutual information maximization (\mimax) and divergence minimization (\dmin) – can generate interesting behaviors in continuous control environments, and provided concrete evaluation metrics for interactive debugging. 
We expect that these approaches, complementary with classical behavior generation techniques through reward engineering would be useful for the long-term goal of creating large data sets of interesting behaviors, and enable algorithmic and architectural breakthrough in RL.
In turn, more diverse data sets of behaviors have the potential to enable more efficient algorithms (e.g. via distillation) for few-shot learning in reinforcement learning, as recently observed in supervised learning.

\begin{ack}
We appreciate feedback and advice from Sergey Levine, Nicolas Heess, Marc Bellemare, Yutaka Matsuo, Pierre Sermanet, Douglas Eck, and Vincent Vanhoucke.
\end{ack}

\small
\bibliographystyle{plainnat}
\bibliography{main}

\newpage
\appendix
\section*{Appendix}
\setcounter{section}{0}
\renewcommand{\thesection}{\Alph{section}}

\section{\braxlines~Benchmark, Hyperparameters, Ablations, and Analyses}
\label{sec:appendix_baselines}

We detail the additional information for experimental results and \braxlines~ Benchmark. While \braxlines~is designed more as a tool to enable rapid creation of environments and behaviors, it also provides minimal quantitative benchmarks that could be utilized as a reference for how to evaluate new environments or new \mimax~and \dmin~algorithms. Such baselines for reward-free algorithms have been lacking, or non-existent, in open-source libraries (see Tables~\ref{tab:reward_max} and~\ref{tab:info_max}). 
\begin{table}[ht]
    \centering
    \begin{tabular}{ |l|c|c|c|c|c|c|c|  }
    \hline
    \hline
     \textbf{Framework} & DDPG & A3C & PPO & TRPO & TD3 & SAC & ES \\
     \hline
     Baselines \cite{baselines} & $\checkmark$ & $\checkmark$ & $\checkmark$ & $\checkmark$ &  &  &\\ \hline
     Stable Baselines \cite{Raffin_Stable_Baselines3_2020} & $\checkmark$ & $\checkmark$ & $\checkmark$ & & $\checkmark$ & $\checkmark$ &\\
     \hline
     RLLab \cite{duan2016benchmarking} & $\checkmark$ & &  & $\checkmark$ & & & $\checkmark$ \\
     \hline
     Tf-Agents \cite{baselines} & $\checkmark$ & & $\checkmark$ & & $\checkmark$ & $\checkmark$ &\\
     \hline
     RLLib \cite{liang2017rllib} & $\checkmark$ & $\checkmark$ & $\checkmark$ & & $\checkmark$ & $\checkmark$ & $\checkmark$\\
     \hline
     PFRL \cite{fujita2021chainerrl} & $\checkmark$ & $\checkmark$ & $\checkmark$ & $\checkmark$ & $\checkmark$ & $\checkmark$ & \\
     \hline
     \textbf{\brax}/\textbf{\braxlines}(ours) & & & $\checkmark$ & & & $\checkmark$ & $\checkmark$ \\
     \hline
     \hline
    \end{tabular}
    \caption{A survey on the support of the most common continuous control, reward maximization algorithms from a set of well-maintained, stable RL baselines.}
    \label{tab:reward_max}
\end{table}

\begin{table}[ht]
    \centering
    \begin{tabular}{ |l|c|c|c|c|c|p{2cm}|  }
     \hline
     \hline
     \textbf{Framework} & \airl & \gail & \gcrl & \diayn & \cdiayn & {\centering Oth. \mimax \newline \citep{hansen2020visr,wardefarley2019discern}}\\
     \hline
     Baselines \cite{baselines} & & $\checkmark$ & & &  & \\ \hline
     Stable Baselines \cite{Raffin_Stable_Baselines3_2020} &  &  &  & & &\\
     \hline
     RLLab \cite{duan2016benchmarking} &  & &  &  & & \\
     \hline
     Tf-Agents \cite{baselines} &  & &  & &  &\\
     \hline
     RLLib \cite{liang2017rllib} &  &  &  & &  &  \\
     \hline
     PFRL \cite{fujita2021chainerrl} &  &  &  & &  & \\
     \hline
     \textbf{\braxlines}(ours) & $\checkmark$ & $\checkmark$ & $\checkmark$ & $\checkmark$ & $\checkmark$ & $\checkmark$  \\
     \hline
     \hline
     \citet{pytorchrl} & & $\checkmark$ &  & & & \\
        \hline
     \citet{eysenbach2019diayn} & & & & $\checkmark$ & & \\
        \hline
     \citet{sharma2020dynamics} & & & & $\checkmark$ & $\checkmark$ & \\
        \hline
     \citet{choi2021variational} & & & & & & \\
      \hline
    \end{tabular}
    \caption{A survey on the support of the most common algorithms in \mimax~and \dmin~families across a set of well-maintained, stable RL baselines. Since standard RL baselines rarely support advanced algorithms, we added a few additional codebases (some from original algorithmic papers). Note that~\citet{choi2021variational} is empty since no code is open-sourced.}
    \label{tab:info_max}
\end{table}

\subsection{Hyperparameters}
\label{sec:hyperparameters}

Hyperparameters for reproducing all the benchmark and experiment results can be found in \url{https://github.com/google/brax/blob/main/brax/experimental/braxlines/experiments}. These experiments can be easily run serially on Colab \url{https://github.com/google/brax/blob/main/notebooks/braxlines/experiment_sweep.ipynb} based on a configuration list of dictionaries such as \url{https://github.com/google/brax/blob/main/brax/experimental/braxlines/experiments/mimax_sweep.py}. Else, users can modify \texttt{run\_experiment()} from \url{https://github.com/google/brax/blob/main/brax/experimental/braxlines/experiments/__init__.py} to launch parallel sweeps on their custom computing clusters.

\paragraph{Experiment Parameters} We used 10 random seeds for each experiment result to compute its mean and variance. Similarly to~\citep{eysenbach2019diayn,sharma2020dynamics,choi2021variational}, we assume prior knowledge on dimensions of interest, e.g. XY-velocities: \texttt{obs\_indices}$=(11,)$ for HalfCheetah, $(13, 14)$ for Ant, and $(22, 23)$ (see \url{https://github.com/google/brax/blob/main/brax/experimental/composer/obs_descs.py}), for learning (except \diaynfull) and evaluation metrics.

\paragraph{PPO Parameters}We used the same exact hyperparameters for PPO as in task-reward RL examples in \brax~(see \url{https://github.com/google/brax/blob/main/notebooks/training.ipynb} for their original Colab and \url{https://github.com/google/brax/tree/main/brax/experimental/braxlines/experiments/defaults.py} for these PPO hyperparameters), except multiplying the \texttt{num\_timesteps} by 2 to allow longer training than in the single-task setting. Fully-connected MLPs with \texttt{linen.swish} activation function and hidden sizes of $[32, 32, 32, 32]$ are used for policy functions, and hidden sizes of $[256, 256, 256, 256, 256]$ for value functions (see \url{https://github.com/google/brax/blob/main/brax/training/networks.py}).

\paragraph{\mimax~Parameters} The hyperparameters are specified in \url{https://github.com/google/brax/blob/main/brax/experimental/braxlines/experiments/mimax_sweep.py}. For discrete latent experiments (\diayn and \diaynfull), number of skills for $z$ \texttt{diayn\_num\_skills} is set to $8$. For continuous latent experiments (\cdiayn), the $z$ dimension is set to $2$. Fully-connected MLPs with \texttt{linen.swish} activation function and hidden sizes of $[32, 32]$ are used for discriminator functions $q(z|\feat(s))$. See Section~\ref{sec:ablations} for ablations on key hyperparameters such as \texttt{spectral\_norm} and \texttt{diayn\_num\_skills}.

\paragraph{\dmin~Parameters} The hyperparameters are specified in \url{https://github.com/google/brax/blob/main/brax/experimental/braxlines/experiments/dmin_sweep.py}. Fully-connected MLPs with \texttt{linen.swish} activation function and hidden sizes of $[32, 32]$ are used for binary discriminator functions. See Section~\ref{sec:ablations} for ablations on key hyperparameters such as \texttt{gradient\_penalty\_weight}.

\paragraph{\composer~Parameters} \braxlines~\composer~is designed as a toolkit for efficiently reusing environment components, observation definitions, and reward functions for composing new tasks or reward-free environments. Example ``hyperparameters'' of these environments are listed in~\url{https://github.com/google/brax/tree/main/brax/experimental/composer/env_descs.py}. Similar to the rest of experiments, hyperparameter sweeps are supported. If the comparison is about the optimizability by PPO with respect to a task score, it is important to ensure that \texttt{score\_fns} are the same across all swept variants. Examples include:
\begin{itemize}
    \item \textbf{Sweeping Reward Function Parameters}: When you have a sum of multiple reward terms, it's helpful to sweep over the scaling parameters to find the optimal trade-off among these to have the best task score of interest \url{https://github.com/google/brax/blob/main/brax/experimental/braxlines/experiments/ant_push_sweep.py}
    \item \textbf{Sweeping Morphology Parameters}: You could parameterize a component to have varying morphologies (e.g. different number of leggs for Ant) \url{https://github.com/google/brax/blob/main/brax/experimental/braxlines/experiments/ant_run_morphology_sweep.py}
\end{itemize}

\subsection{Additional Benchmark Results}
\label{sec:additional_results}

\begin{figure*}[t]
\centering
\includegraphics[width=\textwidth]{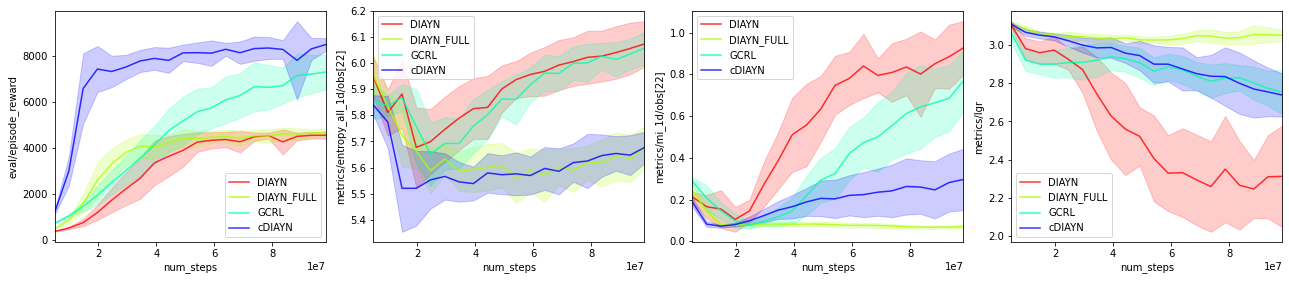}
\vskip -0.05in
\caption[Caption without FN]{
\mimax~results on Humanoid averaged over 10 seeds: (from left to right) (a) episodic discriminator reward, (b, c) \mizs and \entropyall estimates from Algorithm~\ref{alg:mi}, and (d) LGR from Algorithm~\ref{alg:lgr}. Unlike Ant results in~\autoref{fig:mi_max}, \diayn~performs the best both in terms of \mizs and LGR metrics, better than \gcrl. Since \entropyall is the same for \diayn~and \gcrl, this means that \diayn~ acquired better on average controllability/consistency with respect to each given target goal (reasonable since \diayn~is only learning 8 targets, while \gcrl~is learning a infinite set of targets).
}\label{fig:mi_max_humanoid}
\end{figure*}

\begin{figure*}[t]
\centering
\includegraphics[width=\textwidth/3]{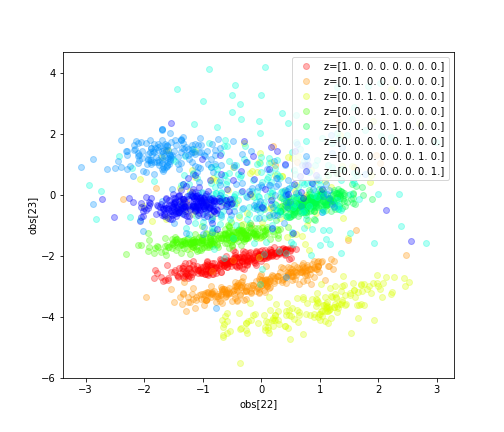}
\vskip -0.05in
\caption{
\mimax~\diayn~skill visualization on Humanoid. Humanoid successfully learns to move in multiple directions. See behavior videos in~\url{https://github.com/google/brax/blob/main/brax/experimental/braxlines}.
}\label{fig:mi_max_humanoid_skills}
\end{figure*}

\paragraph{\mimax} \autoref{fig:mi_max_humanoid} and~\autoref{fig:mi_max_humanoid_skills} show learning curves and skill visualizations for Humanoid respectively. To the best of our knowledge, DADS~\citep{sharma2020dynamics} is one of the few unsupervised RL algorithms that scaled to Humanoid, and we are the first to report successful \diayn~results on it.  

\begin{figure*}[t]
\centering
     \begin{subfigure}[]{0.45\textwidth}
        \centering
         \includegraphics[width=0.99\linewidth]{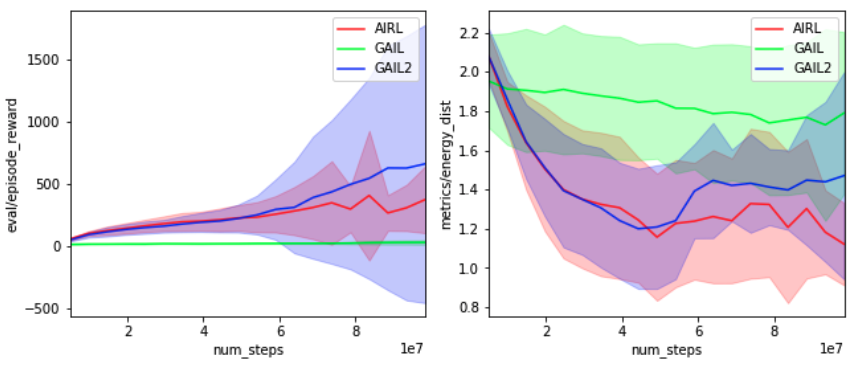}
    \end{subfigure}
         \begin{subfigure}[]{0.45\textwidth}
        \centering
          \includegraphics[width=0.99\linewidth]{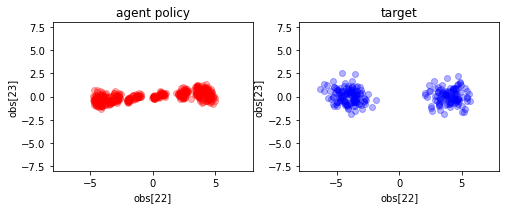}
    \end{subfigure}
\vskip -0.05in
\caption{\dmin~results for Humanoid averaged over 10 seeds: (from left to right) (a, b) episode reward and energy distance metric across training iterations; (c, d) visualizations of learned policy state marginal distribution and the target. Unlike Ant results in~\autoref{fig:d_min}, \airl~and \gail2~significantly outperforms \gail. See behavior videos in~\url{https://github.com/google/brax/blob/main/brax/experimental/braxlines}.
\label{fig:d_min_humanoid}}
\end{figure*}

\paragraph{\dmin} \autoref{fig:d_min_humanoid} shows learning curves and result visualization for Humanoid. Humanoid successfully learns to match the given bi-modal target distribution. 

\begin{figure}[ht]
    \centering
    \begin{subfigure}{0.49\textwidth}
        \centering
        \includegraphics[width=\textwidth]{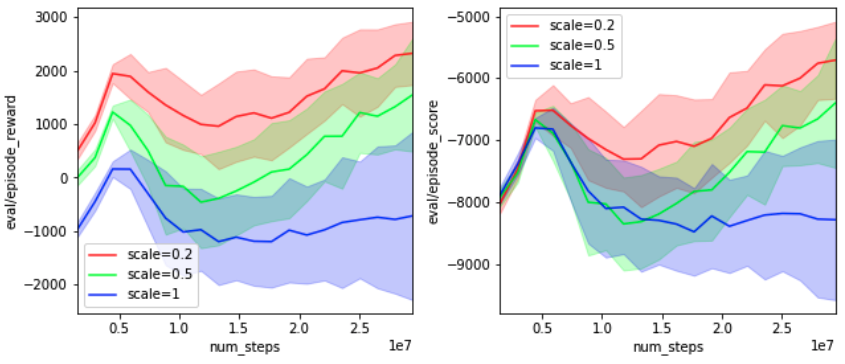}
    \end{subfigure}
    \begin{subfigure}{0.245\textwidth}
        \centering
        \includegraphics[width=\textwidth]{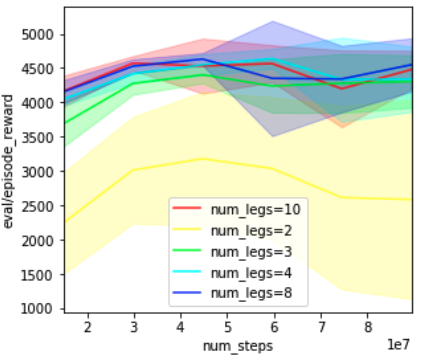}
    \end{subfigure}
   \caption{Task parameter sweeps on (left and middle) Ant Push and (right) morphological Ant Run in~\autoref{fig:composer}. See behavior videos in \url{https://github.com/google/brax/blob/main/brax/experimental/braxlines}.}
    \label{fig:composer_results}
\end{figure}

\paragraph{\braxlines~\composer} \autoref{fig:composer_results} show training results for environment examples in~\autoref{fig:composer}. For Ant Push, we observe that ``scale'' which balances between object-to-target-velocity and ant-to-object reward terms has significant effect on final scores (since ``episode\_reward'' is a variable here, we defined ``episode\_score'' as a consistent metric for evaluation).  For the morphologically-varying Ant Run, we observe that the running performance is largely consistent, with an exception in the case of two legs. Importantly, as described by simple examples in Appendix~\ref{sec:hyperparameters} and~\ref{sec:composer_api_example}, \composer~allows simple programming of environment variations and hyperparameter sweeps. 

\subsection{Ablation Studies and Analysis}
\label{sec:ablations}

This section mainly lists a few design parameters that are critical to getting interesting emergent behaviors. For more examples,  check out our codebase or directly try out our interactive Google Colabs in~\url{https://github.com/google/brax/tree/main/notebooks/braxlines}.

\begin{figure}[ht]
    \centering
       \includegraphics[width=\textwidth]{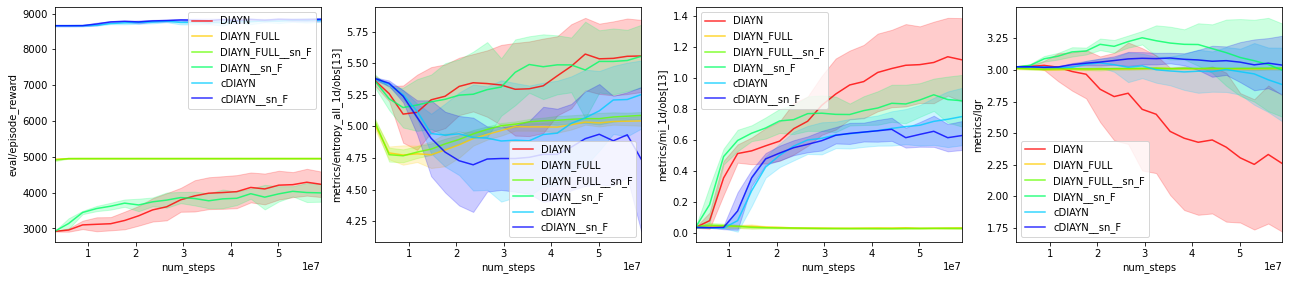}
    \caption{
    \mimax~results on Ant with and without Spectral Normalization (``sn\_F'' means no spectral norm) averaged over 10 seeds: (from left to right) (a) episodic discriminator reward, (b, c) \mizs and \entropyall estimates from Algorithm~\ref{alg:mi}, and (d) LGR from Algorithm~\ref{alg:lgr}. }
    \label{fig:mimax_ant_sn}
\end{figure}

\begin{figure}[ht]
    \centering
       \includegraphics[width=\textwidth]{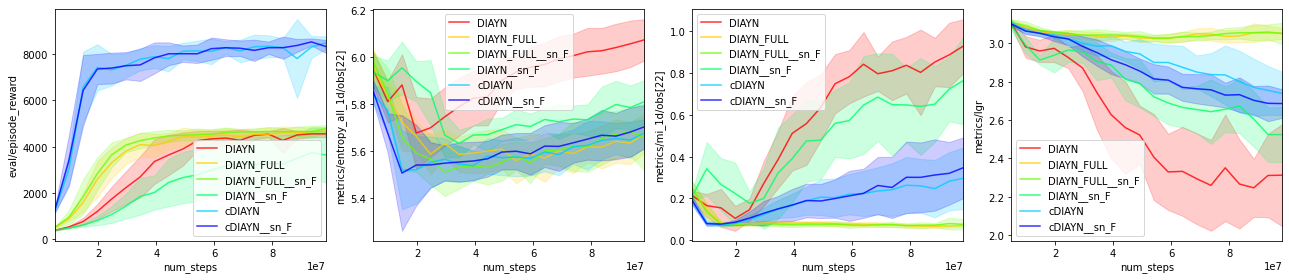}
    \caption{
    \mimax~results on Humanoid with and without Spectral Normalization (``sn\_F'' means no spectral norm) averaged over 10 seeds: (from left to right) (a) episodic discriminator reward, (b, c) \mizs and \entropyall estimates from Algorithm~\ref{alg:mi}, and (d) LGR from Algorithm~\ref{alg:lgr}. }
    \label{fig:mimax_humanoid_sn}
\end{figure}

\begin{figure}[ht]
    \centering
       \includegraphics[width=\textwidth]{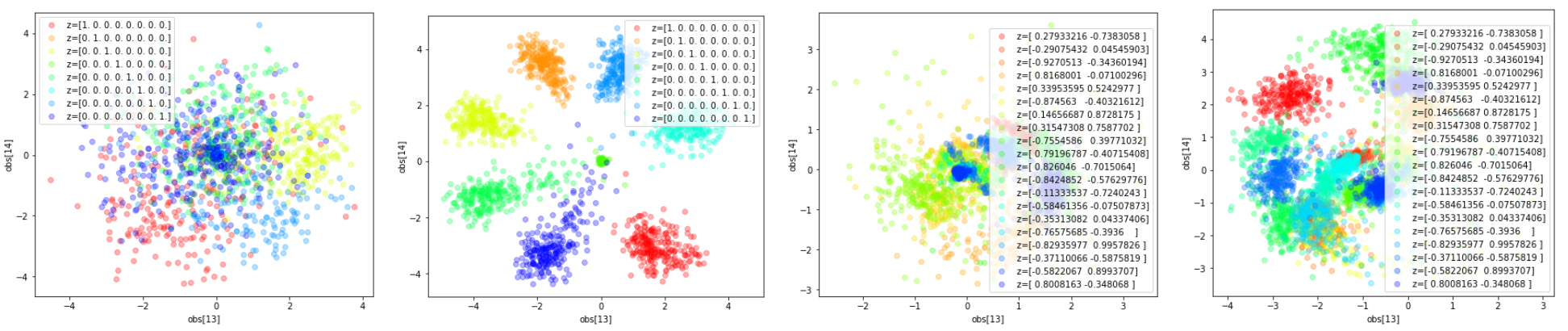}
    \caption{Skills visualization for (from left to right) (a) \diayn~without spectral normalization (SN), (b) \diayn~with SN, (c) continuous \diayn~(\cdiayn) without SN, and (d) \cdiayn~with SN. \diayn~with continuous $z$ has been notoriously difficult even with feature engineering, where~\citet{eysenbach2019diayn} and \citet{sharma2020dynamics} have not showed successful skill learning. A simple combination of PPO~\citep{Schulman2017PPO} and spectral normalization (SN)~\citep{miyato2018spectral} has proven to be surprisingly effective and enabled training a continuous-latent \diayn~(while results are slightly worse than those of discrete-latent \diayn~according to~\autoref{fig:mi_max}.}
    \label{fig:sn}
\end{figure}
\paragraph{Spectral Normalization} \mimax~algorithms exhibit complex learning dynamics through continuously changing reward functions. \citet{choi2021variational} showed that the use of spectral normalization (SN) can significantly stabilize learning and quality of discovered skills. \autoref{fig:mimax_ant_sn} and~\autoref{fig:mimax_humanoid_sn} show that SN does improve \diayn~results substantially in \mizs and \entropyall, in both Ant and Humanoid.
We also show in~\autoref{fig:sn} that even \cdiayn~can discover meaningful skills, which has been difficult in prior works~\citep{eysenbach2019diayn,sharma2020dynamics,choi2021variational}. This result, along with our success with Humanoid, is likely because most prior results built on SAC which is more sample-efficient but less stable than PPO.

\begin{figure*}[t]
\centering
     \begin{subfigure}[]{0.45\textwidth}
        \centering
         \includegraphics[width=0.99\linewidth]{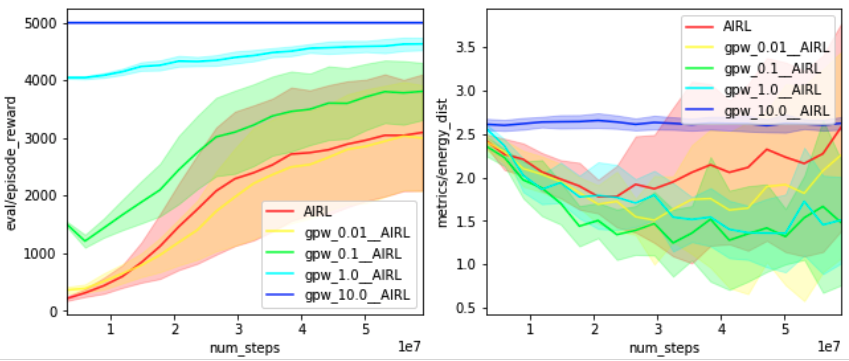}
    \end{subfigure}
         \begin{subfigure}[]{0.45\textwidth}
        \centering
          \includegraphics[width=0.99\linewidth]{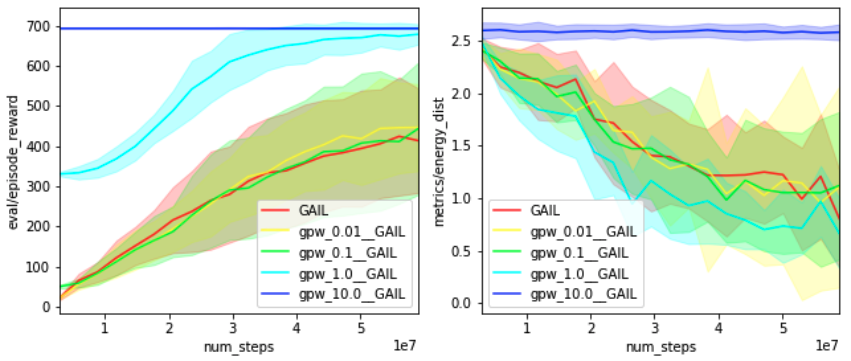}
    \end{subfigure}
\vskip -0.05in
\caption{\dmin~results for Ant with or without Gradient Penalty (``gpw'' stands for gradient penalty weight) averaged over 10 seeds: (from left to right) (a, b) episode reward and energy distance metric across training iterations for \airl; (c, d) episode reward and energy distance metric across training iterations for \gail. Both \braxlines~\airl~and \gail~are stable with or without gradient penalties, and too much penalty can degrade performances.
\label{fig:d_min_gpw}}
\end{figure*}

\paragraph{Gradient Penalty} \dmin~algorithms also exhibit difficult learning dynamics, and many prior works utilized numerous additional implementations to stabilize learning, such as separately learning terminal reward for bias correction~\citep{kostrikov2018discriminatoractorcritic,orsini2021matters} and fictitious play~\citep{lee2020efficient}. Contrary to prior wisdom, \autoref{fig:d_min_gpw} shows that gradient penalty does not significantly improve the learning performances for these bi-modal target matching tasks in Ant or Humanoid.

\begin{figure}[ht]
    \centering
    \begin{subfigure}[]{0.32\textwidth}
        \centering
        \includegraphics[width=\textwidth]{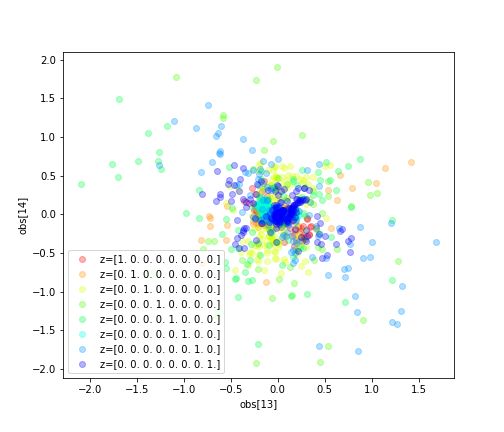}
        \caption{\diaynfull~(full state)}
    \end{subfigure}
    \begin{subfigure}[]{0.32\textwidth}
        \centering
        \includegraphics[width=\textwidth]{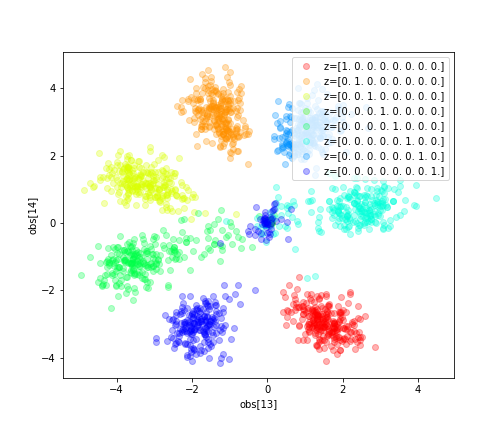}
        \caption{\diayn~(xy)}
    \end{subfigure}
    \begin{subfigure}[]{0.32\textwidth}
        \centering
        \includegraphics[width=\textwidth]{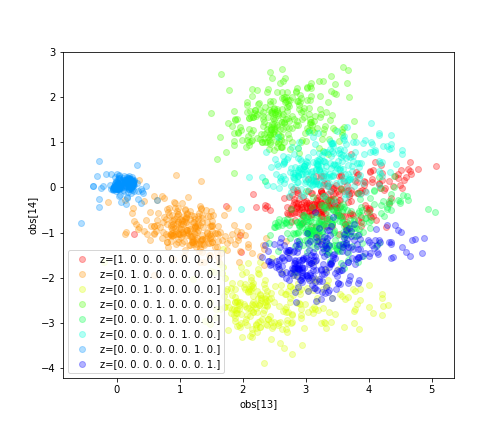}
        \caption{\diayn~(xy) + Run Reward}
    \end{subfigure}
    \caption{The importance of feature engineering (\feat(s)) and reward augmentation (\texttt{env\_reward\_multiplier}) to discover diverse behaviors when running (a) \diayn~\cite{eysenbach2019diayn} with full observation (\diaynfull), (b) \diayn~with XY-velocities (dim=(13,14)), (c) \diayn~with XY + adding original reward on Brax's Ant \cite{brax2021github}. Naively using full states does not lead to interesting behaviors like walking or hopping (it just learns to pose differently). With simple specification of dimensions in observation, however, its learns very effectively. Simple feature engineering  and \mimax~is therefore a powerful tool for behavior engineering without much human efforts. Additionally, if there is some task reward available, it can be added to automatically discover multiple distinguishable ways of accomplishing the same task.}
    \label{fig:feat_eng}
\end{figure}

\paragraph{Feature Engineering $\feat(s)$} \autoref{fig:feat_eng} shows qualitative results of \mimax~with or without feature engineering $o(s)$ (\diayn~or \diaynfull). As different dimensions have different sensitivity to the agent's actions, often the easiest dimension is diversified first to maximize the objective in discriminative \mimax, i.e. in \diaynfull~Ant just learns to make slightly different poses. By removing uninteresting and trivial dimensions for control, the agent can effectively focus on controlling the key dimensions such as XY velocities (dim$=(13,14)$). Our fast interactive tool allows efficiently iterating different feature choices.

\begin{figure}[ht]
    \centering
    \includegraphics[width=\textwidth]{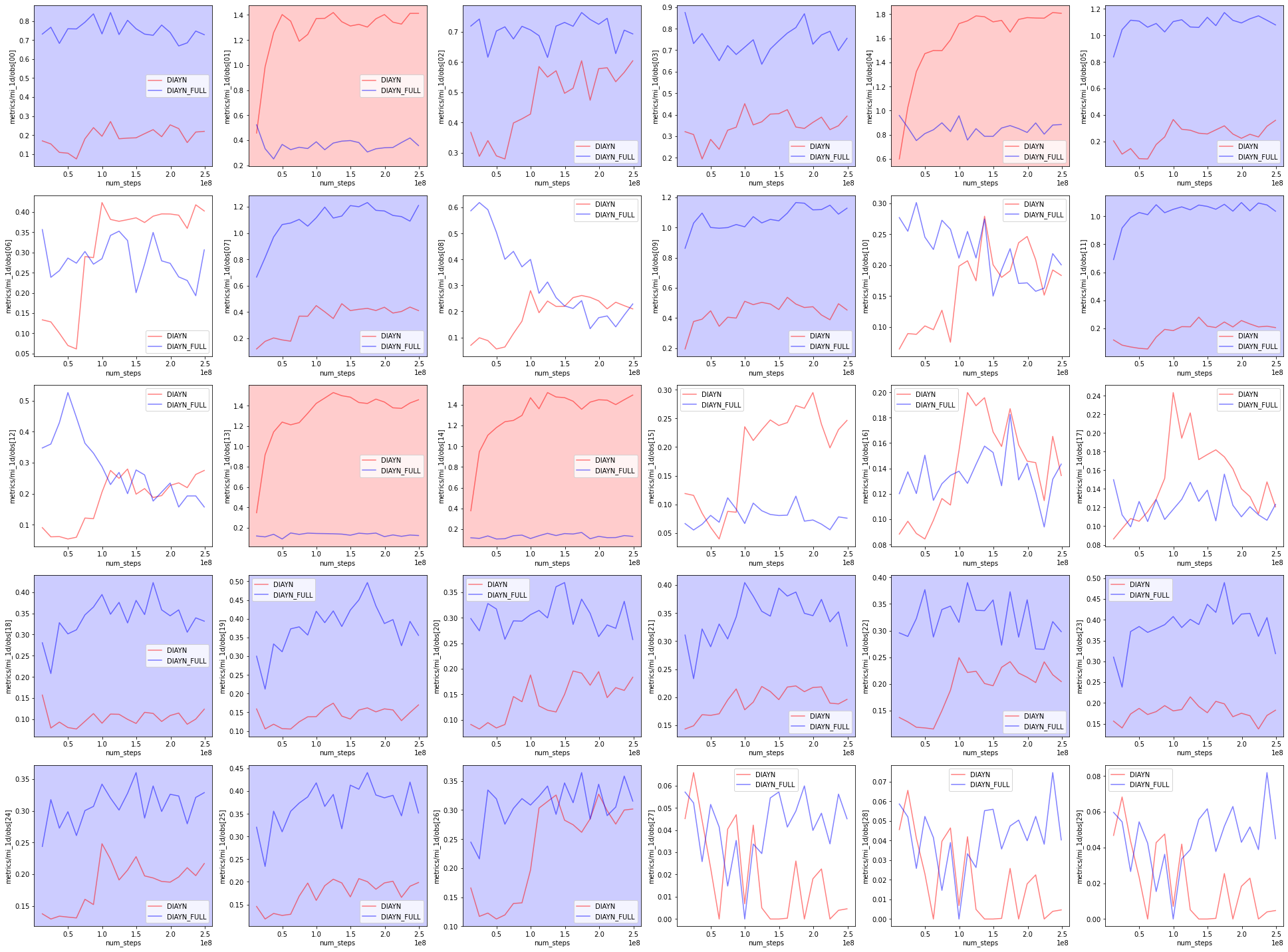}
    \caption{Mutual information evaluations across first 30 dimensions on Ant. We analyzed two configurations: \diayn~ uses $\feat(s[13,14])$ (i.e., the XY velocities of the root), while \diaynfull~ uses the full state. The red background indicates when \diayn~ is higher than \diaynfull~ more than 80\% of the time and blue if vice versa. While it is unsurprising that \diayn~excels at gaining controllability over specified dimensions (13, 14), it is interesting that \diayn~ additionally gained controllability over dim=1 (i.e., the first rotation angle of the root) and lost controllability over dim=(0, 2) (the height and other rotation angles of the root) as an indirect a consequence of empowering dim=(13,14).}
    \label{fig:mi_ant_full}
\end{figure}

\begin{figure}[ht]
    \centering
    \includegraphics[width=\textwidth]{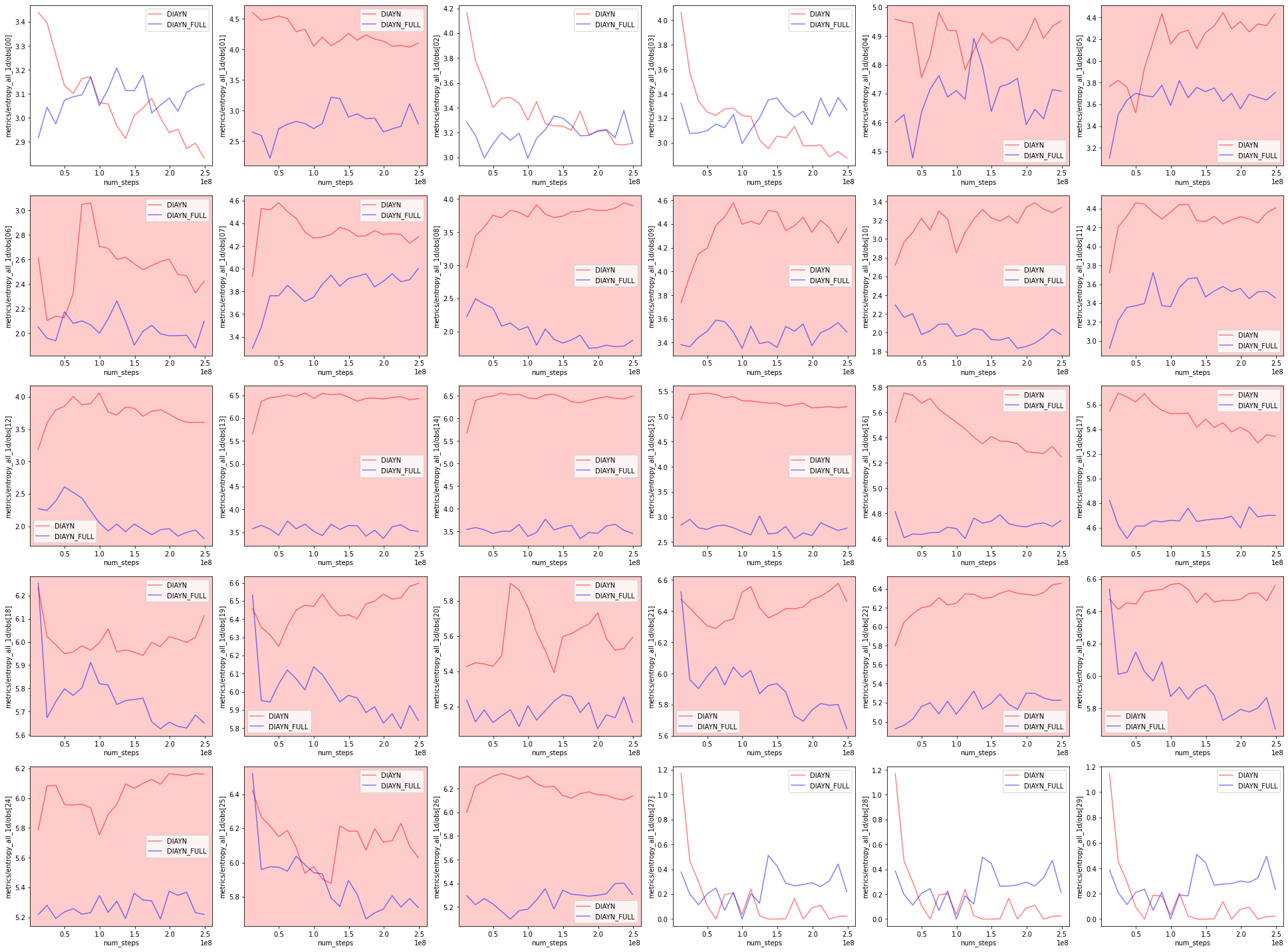}
    \caption{Entropy evaluations across first 30 dimensions on Ant. \diayn~ uses $\feat(s[13,14])$ (corresponding to xy velocities), while \diaynfull~uses identity  $\feat$ (full observation state). It is noticeably to note that in all dimensions \diayn~(XY) gains higher state entropies than \diayn~(full state). Contrasting this result with~\autoref{fig:mi_ant_full}, we can conclude that while \diayn~(XY) can thoroughly increase marginal entropies, it can only gain controllabilities (low conditional entropies) over a limited set of the dimensions. In cases where you only care about marginal entropies such as for risk-seeking exploration, this may be desired over controllability.}
    \label{fig:ent_ant_full}
\end{figure}

Additionally, \autoref{fig:mi_ant_full} and~\autoref{fig:ent_ant_full} provide a deeper look into MI and $H(s)$ estimates across first 30 dimensions Ant, \diayn~or \diaynfull. This essentially evaluates intrinsic controllability that the agent has over the environment, from which we can gain quantitative intuitions about the agent's behaviors as well as its affinity with the environment. For example, \diayn learns to run in different directions (dim=(13,14)), but as it runs it does not change its Z-axis orientation (dim=1) and therefore indirectly gained controllability over it. If we can measure such indirect relationships among dimensions, we can more efficiently design minimal features $\feat(s)$. Furthermore, combining with the insights from policy information capacity (PIC)~\citep{furuta2021pic}, we can choose to design environments or random initial policies such that the initial MI is already high with respect to the dimensions of interest, i.e. designing environments and agents for optimizability~\citep{furuta2021pic,reda2020learning}.

\paragraph{Task-Reward Reward Augmentation} When a task reward is combined with \mimax~ (\texttt{env\_reward\_multiplier} $>0$), it essentially diversifies policies \textit{while} accomplishing a given task, i.e. learning multiple independent ways of accomplishing the task. Such auxiliary task reward can sometimes be as simple as a positive survival reward (which is used in all our \mimax~and \dmin~experiments as listed in~\autoref{tab:alg_summary}), if the environment itself is non-trivial and has termination like the Humanoid environment~\citep{sharma2020dynamics}. In~\autoref{fig:feat_eng}, the standard Ant's forward running reward is combined, and it learns different gait behaviors.  

\clearpage
\section{Latent Goal Reaching}
\label{sec:lgr}
Algorithm~\ref{alg:lgr} estimates the LGR metric~\citep{choi2021variational} of a policy $\pi(a|s,z)$. This has exact correspondence with the standard goal-reaching objective in \gcrl.

\begin{algorithm}[H]
\SetAlgoLined
\KwIn{agent $\pi(a|s, z)$; $\feat(s)$; goals $\feat^*_{1:L}=\feat(s^*_{1:L})\sim p^*(s)$ ; inference $q(z|\feat(s))$}
\KwOut{LGR estimate $\hat{\text{LGR}}$} 
  Sample $z_{l,n} \sim   q(z|s^*_l)$   \blue{// Infer $N$ latent intents per goal $s^*_l$} \\
    Sample $s_{l,n,m,t}\sim\pi(\cdot|\cdot, z_{l,n})$ in $\mu$ \blue{// Collect $M$ episodes of horizon $T$ per $z_{l,n}$}  \\
  \blue{// Compute average goal-reaching performance per $z_{l,n}$ for recovering $s^*_l$}  \\
  $\hat{\text{LGR}} = \sum_{l,n, m,t} -(\feat^*_l-\feat(s_{l,n,m,t}))^2/\sigma^2$ \\
 \caption{Latent Goal Reaching~\citep{choi2021variational}}
  \label{alg:lgr}
\end{algorithm}

\subsection{Justification for Negative Latent Goal Reaching as An Empowerment Approximation}
\label{sec:lgr_deriv}
Given $p^*(z) = \arg\max_{p(z)}I_\pi(z, \feat({S^\mu}))$, the empowerment per environment $\mu$ is given by:
\begin{align}
   p^*(z)p(\feat({S^\mu})|z)&= p^*(\feat({S^\mu}))p^*(z|\feat({S^\mu}))\label{eq:jointstar}\\
    I_\pi^*(z, \feat({S^\mu})) &=  H^*(\feat({S^\mu})) - H^*(\feat({S^\mu})|z) \nonumber\\
    &= H^*(\feat({S^\mu})) + \mathbb{E}_{ p^*(\feat({S^\mu}))p^*(z|\feat({S^\mu}))}\left[ \log p(\feat({S^\mu})|z)\right] \label{eq:istar}
\end{align}
Importantly, in joint distributions in Eq.~\ref{eq:jointstar}, $p(\feat({S^\mu})|z)$ does not depend on the choice of $p(z)$. We then make two important assumptions: (1) we are given with samples of $p^*(\feat({S^\mu}))$ and $p^*(\feat({S^\mu}))$ is fixed \textit{regardless of the choice of agent}, and (2) we are given with (or could train models to approximate) $q(z|\feat({S^\mu}))\approx p^*(z|\feat({S^\mu}))$. While both assumptions are difficult in practice to satisfy accurately, since each depends on $p^*(z)$ which clearly depends on the agent policy $\pi(a|o,z)$, they do allow deriving a connection between generalized goal-reaching objective and empowerment estimation. In particular, Eq.~\ref{eq:istar} can now be estimated by:    
\begin{align}
    I_\pi^*(z, S^\mu)
    &\approx \mathbb{E}_{ p^*(\feat({S^\mu}))q(z|\feat({S^\mu}))}\left[ \log p(\feat({S^\mu})|z)\right] + C_1,\label{eq:istarapprox}
\end{align}
where $C_1=H^*(\feat({S^\mu}))$ is now just a constant offset due to our assumption (1) that does not depend on the evaluation agent $\pi$. Eq.~\ref{eq:istarapprox} is equivalently the variational autoencoder objective~\citep{kingma2014autoencoding}, except in RL we need to somehow estimate $\log p(\feat({S^\mu})|z)$. One way is to fit a density estimation model~\citep{lee2020efficient,berseth2021smirl}, but to make a direct connection to goal-reaching performances, we instead use a non-parametric density estimation based on a mixture of Gaussians. Following the notations in Algorithm~\ref{alg:lgr}, 
\begin{align}
  \log p(\feat^*_l|z_{l,n}) &\approx \log \frac{1}{MT}\sum_{m,t}\mathcal{N}(\feat^*_l|\feat_{l,n,m,t}, \sigma^2 I)\\
    & \geq \sum_{m,t} -(\feat^*_l-\feat_{l,n,m,t})^2/\sigma^2 + C_2,\label{eq:logqstarapprox}
\end{align}
where Eq.~\ref{eq:logqstarapprox} follows from a simple application of Jensen's inequality, and $C_2$ is again a constant offset that does not depend on $\pi$.

\clearpage
\section{Prior Evaluation Methodologies for \mimax~and \dmin~Algorithms}
\label{sec:prior_evaluation}
In this section, we briefly summarize prior evaluation methodologies for \mimax~and \dmin~families, and relation to the metrics in the \braxlines.
Since the inferred reward function dynamically changes during training, could have an arbitrary range of values, and largely depends on the different implementation and hyperparameter choices such as $z$-space in \mimax, or reward transformation in \dmin, it is difficult to apply episodic reward for quantitatively evaluating convergence or final performances.
As in \autoref{tab:prior_evaluation}, most of evaluations for \mimax~and \dmin~have relied on qualitative comparisons, visualization of rollout videos or density plots~\citep{eysenbach2019diayn, sharma2020dynamics, ghasemipour2020divergence, lee2020efficient}, or down-stream task performances~\citep{eysenbach2019diayn,sharma2020dynamics,sharma2020skillsrobotics}. Qualitative evaluation might be hard to ensure objectiveness and automate the whole process, and the evaluation with down-stream tasks may not correspond with the quality of generated behaviors themselves.
While previous \dmin~algorithms mostly focus on imitation learning and frequently use task rewards directly to compare how close to the experts they got~\citep{ho2016generative,fu2018learning,ghasemipour2020divergence}, such evaluation is not suitable for the distribution sketching case in our paper.

Recently, \citet{kim2021unsupervised} employ SEPIN@$k$ and W SEPIN, which are proposed in the disentangled representation learning literature~\citep{Do2020Theory}, for the quantitative evaluation of \mimax~algorithms.
Both SEPIN@$k$ and W SEPIN consider the mutual information between the skills $Z$ and the last states of the trajectories with specified dimensions $\feat(S_{T})$. SEPIN@$k$ is the top-$k$ average of $\text{MI}(\feat(S_{T}), Z_i ; Z_{\neq i})$ over skills $i=1, 2, \dots$, and W SEPIN is the weighted sum of $\text{MI}(\feat(S_{T}), Z_i ; Z_{\neq i})$ over skills. They also propose to simply use $\text{MI}(\feat(S_{T}), Z)$ as a information theoretic metrics. See \citet{kim2021unsupervised} for the detailed discussion. Because they consider the last states the agents reached and measure how they are different depending on $z$, these might be similar metrics to LGR.

\begin{table}[ht]
\small
\centering
\scalebox{0.9}{
\begin{tabular}{|c|l|c|c|c|}
\hline
\hline
\textbf{Family} & \textbf{Metrics} & \textbf{Type} & \textbf{Hyperparameter-Agnostic} & \textbf{References} \\
\hline
    \mimax & Episodic Discriminator Reward & Quantitative & No & --\\
    \mimax & Diversity in Rollout Videos & Qualitative & Yes & \citep{eysenbach2019diayn, sharma2020dynamics, florensa2017stochastichrl,kim2021unsupervised} \\
    \mimax & Downstream Task Performance & Quantitative & Yes & \citep{eysenbach2019diayn,sharma2020dynamics,florensa2017stochastichrl,hansen2020visr,gregor2016variational} \\
    \mimax & SEPIN@$k$, W SEPIN~\citep{Do2020Theory} & Quantitative & Yes & \citep{kim2021unsupervised}\\
    \dmin & Episodic Inferred Reward & Quantitative & No & --\\
    \dmin & Density Visualization & Qualitative & Yes & \citep{ghasemipour2020divergence, lee2020efficient} \\
    \dmin & Imitation Task Performance & Quantitative & Yes & \citep{ho2016generative,fu2018learning,ghasemipour2020divergence} \\
\hline
    \mimax & Particle-based MI Estimation & Quantitative & Yes & Ours, \citep{furuta2021pic} \\
    \mimax & Latent Goal Reaching & Quantitative & Yes & Ours, \citep{choi2021variational} \\
    \dmin& Energy Distance & Quantitative & Yes & Ours \\
\hline
\end{tabular}
}
\vspace{0.5pt}
\caption{A summary of prior evaluation methods for \mimax~and \dmin~algorithms.}
\label{tab:prior_evaluation}
\end{table}

\newpage
\section{\braxlines~\composer~API Example}
\label{sec:composer_api_example}

We made possible in \braxlines~\composer~to design continuous control tasks with under 50 lines of code (including comments). The code snippet below shows how to construct a manipulation, goal-conditioned task that we coined \emph{ant-push}. We made available several off-the-shelf components\footnote{Available at \url{https://github.com/google/brax/tree/main/brax/experimental/composer/components}} for rapid prototyping. These components were designed with flexibility in mind, making them amenable for procedurally-generated continuous control tasks or for learning modular policies~\cite{huang2020policy, kurin2021body} that can control multiple morphologies.   

For a complete running example and a more in-depth introduction to \braxlines~\composer, check out the \emph{Composer Basics}\footnote{Available at \url{https://colab.research.google.com/github/google/brax/blob/main/notebooks/composer/composer.ipynb}} notebook.

\begin{minted}[linenos, frame=lines]{python}
from brax.experimental.composer import composer

env_desc = dict(
    components=dict(  # component information
        agent1=dict(
            # pro_ant is a provided off-the-shelf component
            # defined at components/pro_ant.py
            component='pro_ant',  
            # among other things, designers can configure 
            # the ant morphology
            component_params=dict(num_legs=6) # ant with 6 legs
        ),  
        cap1=dict(
            # the singleton is the ball the ant controls
            # defined in components/singleton.py
            component='singleton',  
            component_params=dict(size=0.5), # a ball with radius 0.5
            pos=(1, 0, 0),  # where to place a capsule object
            reward_fns=dict( # reward1: a GCRL task
                goal=dict(  
                    reward_type='root_goal',
                    sdcomp='vel',
                    target_goal=(4, 0, 0) # a target velocity for the object
                )
            )
        )
    ),
    edges=dict(  # edge information
        agent1__cap1=dict(  # edge names use sorted component names
            # observers extract relevant information for control           
            extra_observers=[  # add agent-object position diff as an extra obs
                dict(observer_type='root_vec')
            ],
            reward_fns=dict(  # reward2: make the agent close to the object
                dist=dict(reward_type='root_dist')
            ),
        ),
    )
)
env = composer.create(env_desc=env_desc) # Composer returns a functional environment
\end{minted}    

\end{document}